\documentclass[12pt]{iopart}

\usepackage[utf8]{inputenc}
\usepackage[english]{babel}
\usepackage{graphicx}
\usepackage{amsfonts}
\usepackage{cite}
\usepackage[dvipsnames]{xcolor}
\usepackage{hyperref}
\usepackage{float}
\usepackage{placeins}

\begin{document}
\title[Anomalous Diffusion through Convolutional Transformers]{Characterization of anomalous diffusion through convolutional transformers}

\author{Nicol\'as Firbas$^{1,\dagger}$, \`Oscar Garibo-i-Orts$^2$, Miguel \'Angel Garcia-March$^3$, J. Alberto Conejero$^3$}
\address{$^1$ DBS - Department of Biological Sciences, National University of Singapore, 16 Science Drive 4, Singapore 117558, Singapore.} 
\address{$^2$VRAIN -  Valencian Research Institute for Artificial Intelligence,  Universitat Polit\`ecnica de Val\`encia, 46022 Val\`encia, Spain.}
\address{$^3$IUMPA - Instituto Universitario de Matem\'atica Pura y Aplicada, Universitat Polit\`ecnica de Val\`encia, 46022 Val\`encia, Spain.}
\ead{$\dagger$ aconejero@upv.es (Corresponding author).}
\vspace{10pt}
\begin{indented}
\item[]\today
\end{indented}

\begin{abstract}

The results of the Anomalous Diffusion Challenge (AnDi Challenge) \cite{munoz-gil2021_AnDi_Challenge} have shown that machine learning methods can outperform classical statistical methodology at the characterization of anomalous diffusion in both the inference of the anomalous diffusion exponent $\alpha$ associated with each trajectory (Task 1), and the determination of the underlying diffusive regime which produced such trajectories (Task 2). Furthermore, of the five teams that finished in the top three across both tasks of the AnDi challenge, three of those teams used \textit{recurrent neural networks} (RNNs). While RNNs, like the \textit{long short-term memory} (LSTM) network, are effective at learning long-term dependencies in sequential data, their key disadvantage is that they must be trained sequentially. In order to facilitate training with larger data sets, by training in parallel, we propose a new \textit{transformer} based neural network architecture for the characterization of anomalous diffusion. Our new architecture, the \textit{Convolutional Transformer} (ConvTransformer) uses a bi-layered convolutional neural network to extract features from our diffusive trajectories that can be thought of as being words in a sentence. These features are then fed to two transformer encoding blocks that perform either regression (Task 1) or classification (Task 2). To our knowledge, this is the first time transformers have been used for characterizing anomalous diffusion. Moreover, this may be the first time that a transformer encoding block has been used with a convolutional neural network and without the need for a transformer decoding block or positional encoding. Apart from being able to train in parallel, we show that the ConvTransformer is able to outperform the previous state of the art at determining the underlying diffusive regime (Task2) in short trajectories (length 10-50 steps), which are the most important for experimental researchers. 

\end{abstract}

\section{Introduction}
It could be said that the study of diffusion began in 1827 when Brown first observed the motion, which now carries his namesake, of pollen from \textit{Clarkia pulchella} suspended in water \cite{Brown-1827}. This movement results from small particles being bombarded by the molecules of the liquid in which they are suspended, as was first conjectured by Einstein and later verified by Perrin \cite{perrin1909movement}. Though Brown never managed to explain the movement he observed, we now know that Brownian motion is a kind of normal diffusion.\medskip

To describe diffusion, we can consider the following analogy: Let us imagine a particle being an ant, or some other diminutive explorer,  we can then think of mean squared displacement (MSD), which can be written as $\langle \bold{x}^2\rangle$, as the portion of the system that it has explored. For normal diffusion such as Brownian motion, the relation between the portion of explored region and time is linear, $\langle\bold{x}^2\rangle \sim t$. As time progresses, the expected value of distance explored by our ant (MSD) will remain constant. In contrast to normal diffusion, anomalous diffusion is characterized by $\langle \mathbf{x}^2 \rangle \sim t^\alpha, \alpha \neq 1$. Anomalous diffusion can be further subdivided into super-diffusion and sub-diffusion, when $\alpha >1 $ or $\alpha < 1$, respectively. To continue using the analogy of our ant, an intuitive example of sub-diffusion would be diffusion on a fractal. In this case, it is easy to see how, as time progresses and our ant ventures into zones of increasing complexity, its movement will in turn be slowed. Thus the relationship of space explored and time will be $\langle \bold{x}^2 \rangle \sim t^\alpha, \alpha < 1$. Conversely, if we give our ant wings and have it randomly take flight at random times $t_i$ sampled from $t^{-\sigma-1}$ with flight times positively correlated to the wait time, then for $\sigma \in (0,2)$ we would have a super-diffusive L\'evy flight trajectory.\medskip

Since the discovery of Brownian motion, many systems have shown diffusive behavior that deviates from the normal one, where MSD scales linearly with time. These systems can range from the atomic scale to complex organisms such as birds. Examples of such diffusive systems include ultra-cold atoms \cite{Sagi2012-ultra-cold-atoms}, telomeres in the nuclei of cells \cite{Bronstein2009-telomere-diffusion}, moisture transport in cement-based materials, the free movement of arthropods \cite{Nagaya2017_pill_bug_diffusion}, and the migration patterns of birds \cite{Vilk20201_stork_migration_diffusion}. Anomalous diffusive patterns can even be observed in signals that are not directly related to movement, such as heartbeat intervals and DNA \cite[pg. 49-89]{bunde_1994-fractals-in-science}. The interdisciplinary scope of anomalous diffusion highlights the need for modeling frameworks that are able to quickly and accurately characterize diffusion in real-life scenarios, where data is often limited and noisy.\medskip

Despite the importance of anomalous diffusion in many fields of study \cite{2015ManzoROPP}, detection and characterization remain difficult to this day. Traditionally, mean squared displacement (MSD$(t) \sim t^\alpha$) and its anomalous diffusion exponent $\alpha$ have been used to characterize diffusion. In practice, computation of MSD is often challenging as we often work with a limited number of points in the trajectories, which may be short and/or noisy, highlighting a need for a robust method for real-world conditions. The problem with using $\alpha$ to characterize anomalous diffusion is that trajectories often have the same anomalous diffusion exponent while having different underlying diffusive regimes. An example would be the motion messenger RNA (mRNA) in a living \textit{E. coli} cell. The individual trajectories of the mRNA share roughly the same $\alpha$ despite their trajectories being quite distinct \cite{metzler2014_diffusion_models}.\medskip

Being able to classify trajectories based on their underlying diffusive regime is useful because it can shed light on the underlying behavior of the particles undergoing diffusion. This could be more important for experimental researchers, which may be more concerned with how a particle moves not necessarily how much it has moved. In this vein, the AnDi (Anomalous Diffusion) Challenge organizers identified the following five diffusive models \cite{munoz-gil2021etai}: the \textit{continuous-time random walk} (CTRW) \cite{1975Scher}, \textit{fractional Brownian motion} (FBM) \cite{mandelbrot_vanness_1968}, the \textit{L\'evy walk} (LW) \cite{1994Klafter}, \textit{annealed transient motion} (ATTM) \cite{nonErgodic-subdiffusion}, and \textit{scaled Brownian motion} (SBM) \cite{selfSimilarProcesses-PhysRevE}, with which to classify trajectories. This information is not meant to supplant traditional MSD-based analysis, rather, it is meant to give us additional information about the underlying stochastic process behind the trajectory. For example, for a particular exponent $\alpha$, one may not have access to an ensemble of homogeneous trajectories. Moreover, one cannot assure that all measured trajectories have the same behavior and can therefore be associated with the same anomalous exponent $\alpha$. In these cases, it may be possible to explain the behavior of the diffusing particles by using what we know about five models mentioned above.\medskip

The first applications of machine learning methods to the study of diffusion aimed to discriminate among confined, anomalous, normal qualitatively, and directed motion \cite{dosset2016automatic,kowalek2019classification}. These ML models did not extract quantitative information nor determining did they determine the underlying physical model. At first long short-term memory (LSTM) recurrent neural networks \cite{lstm1997} were considered for the analysis of anomalous diffusion trajectories from experimental data in \cite{bo2019measurement}. Later, Mu\~{n}oz-Gil et al. \cite{MuozGil2020SingleTC} computed the distances between consecutive positions in raw trajectories and normalized them by dividing by the standard deviation. Then, their cumulative sums fed random forest algorithms that permit to infer the anomalous exponent $\alpha$ and to classify the trajectory in one of these models, CTRW, FBM, or LW. Random forests and gradient boosting methods were already considered for 
the study of fractional anomalous diffusion of single-particle trajectories in \cite{Janczura2020Classification,Loch-Olszewska2020Impact}.\medskip

The results of the AnDi Challenge \cite{munoz-gil2021_AnDi_Challenge} showed that machine learning (ML) algorithms outperform traditional statistical techniques in the inference of the anomalous diffusion exponent (Task 1) and in the classification of the underlying diffusion model (Task 2), across one, two, and three dimensions. Some of the most successful techniques consisted of: a couple of convolutional layers combined with some bidirectional LSTM layers and a final dense layer \cite{Garibo2021_AnDi}, two LSTM layers of decreasing size with a final dense layer \cite{Argun2021ClassificationIA}, a WaveNet encoder with LSTM layers \cite{Li_2021} , or the extraction of classical statistics features combined with three deep feed-forward neural networks \cite{Gentili2021_Condor_AnDi_Volpe}.\medskip

As we can see, the best performing methods from the AnDi Challenge were either entirely based on LSTMs recurrent neural networks or incorporated them as part of a larger architecture. For many years, LSTM have been one of the most successful techniques in natural language processing (NLP) and time series analysis. As a matter of fact, Google Translate algorithm is a stack of just seven large LSTM layers \cite{wu2016google}. However, since the landmark paper \textit{Attention is All You Need} \cite{vaswani2017attention}, transformers have become the dominant architecture in NLP, where they have surpassed previous models based on convolutions and recurrent neural networks \cite{wolf2019_huggingface_transformer}. Inspired by the transformers' success and by drawing a parallel between the sequential nature of language and the diffusion of a single particle, we propose a new architecture combining convolutional layers with transformers, that we will call a convolutional transformer: the \textit{Convolutional Transformer} (ConvTransformer) \medskip


\subsection{The Convolutional Transformer}

 The ConvTransformers method has been applied to both the inference the anomalous diffusion exponent $\alpha$ (Task 1) and the determination of the underlying diffusion model (Task 2). As the name suggests, the ConvTransformer uses two convolutional layers followed by a transformer encoding block. However, unlike the transformer in \cite{vaswani2017attention}, our method uses only two transformer encoding blocks in sequence without a transformer decoding block or positional encoding. The convolutional layers behave as an encoder extracting both linear and non-linear features from the trajectory while retaining spatiotemporal awareness eliminating the need for positional encoding. These features are then passed to the transformer encoding layers where attention is performed upon them. The ConvTransformer structure can be intuitively understood if we consider a single trajectory is akin to a sentence. In this analogy the CNNs are used to create pseudo-words, which are the features produced by the CNNs. Finally, we perform attention twice on the pseudo-words with our transformer encoder, which allows us to determine which features are the most important, and from there we are able to obtain either our $\alpha$ or the underlying diffusive regime model.\medskip

The ConvTransformer does not require positional encoding because the CNN kernel moves across the trajectory to create the features. As the CNN kernel moves along the trajectory, it learns positional information, negating the need for positional encoding prior to the transformer encoding block. This was assessed by testing the ConvTransformer on the Task 2 with five-fold validation on a training set of size 50K (32K for training, 8K for validation, and 10K for testing), using the same set of hyper-parameters, with and without the trigonometric encoding scheme used in Vaswani \emph{et al.} 2017\cite{vaswani2017attention}. A five-fold validation of models trained with and without positional encoding showed showed that model classification accuracy decreased with positional encoding to a mean 72.39\% and standard deviation of 4.86 from 75.66\% and standard deviation of 1.54. Thus, positional encoding did not improve ConvTransformer performance, and it was omitted from the model.\medskip 

\begin{figure}[h]
	\centering	
	\includegraphics[width=.7\textwidth]{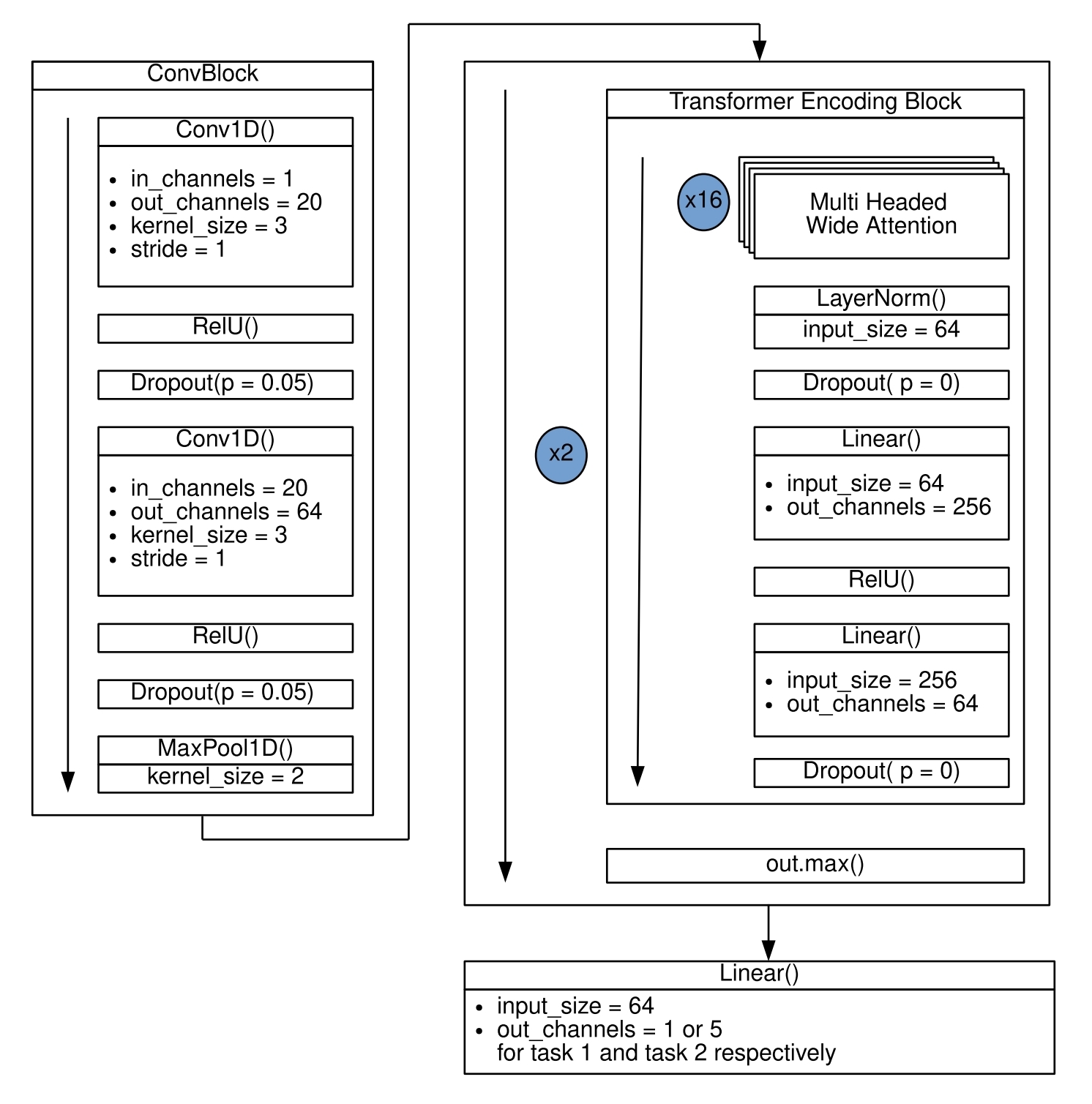}
	\caption{Visual representation of the ConvTransformer structure.}
	\label{fig:ConvTransformer_Diagram}
	\vspace{-1em}
\end{figure}

In Figure \ref{fig:ConvTransformer_Diagram}, we show a diagram detailing the structure of the ConvTransformer. As was previously said, the ConvTransformer uses two convolutional layers, one which scales our trajectory up to 20 features, and a second one takes those 20 features and outputs 64 features. This structure allows the CNN to learn lower-level features first and then refine those features in the subsequent layer. Both convolutional layers use a kernel size of 3, a stride of 1, and are followed by a rectified linear unit function (ReLU), and a dropout with a 5\% probability of setting a learned parameter to 0 to avoid over-fitting. At the end of the convolutional block, we do a pooling with kernel size 2, which cuts the length of our output in half. This helps conserve video memory (VRAM), optimizing resource consumption and democratizing the model as it can run in consumer-grade hardware. The transformer encoding block follows the basic structure of the transformer encoding block from \cite{vaswani2017attention}. It uses wide multi-headed attention with 16 attention heads. The attention mechanism is then followed by layer normalization and dropout. This output feeds two linear layers separated by a ReLU, which ultimately goes to another dropout. This transformer encoding block then feeds into another transformer encoding block. This output of this final transformer encoding block then goes to a max function, which gets the largest value of the output tensor by column. Finally, the output of the ConvTransformer feeds a linear layer that outputs size one for Task 1 or size five for each of the categories in Task 2.

\FloatBarrier

\section{Methods}
\subsection{Generation of Data sets}
All of the data sets used to train and test our models were generated with the Python 3 package provided by the AnDi Challenge \cite{munoz-gil2021_AnDi_Challenge}. The code was made freely available by the AnDi Challenge organizers at: \url{https://github.com/AnDiChallenge/ANDI_datasets}.\medskip

\subsection{Hyper-Parameter Selection}

In order to select the hyper-parameters, we chose a relatively small dataset to train all permutations of these hyper-parameters using five-fold validation to ensure that model performance for a given hyper-parameter set was reliable across runs. The hyper-parameters sets were assessed using a data set with 50K trajectories of lengths [10, 1000] across both Task 1 and Task 2, which was broken up into 32K training, 8K validation, and 10K testing. Then the sets of tested hyper-parameters were evaluated as per the five-fold validation, and the hyper-parameters were chosen for the final model, except the Learning Rate (LR), which has to be scaled up with respect to training set size and batch size \cite{smith2018dontDecreaseLRIncreaseBS}. In the selection process, model performance and feasibility were assessed to ensure that model training could take place on our hardware within a reasonable amount of time. Different sets of hyper-parameters were tested for both Task 1 and Task 2 but, in the end, we found that the same hyper-parameters work well for both tasks. These final hyper-parameters can be found in Table \ref{tab:hyper_parameters}.\medskip

In order to generalize learning rate to larger training data sets, we used the results from \cite{smith2018dontDecreaseLRIncreaseBS} to relate the noise scale (g) during training to batch size (B), training size (N), and learning rate ($\epsilon$), as shown in Equation \ref{eq:Noise_Scale_SGD}.

\begin{equation}
\centering
	g \approx \epsilon \cdot \frac{N}{B}
	\label{eq:Noise_Scale_SGD}
\end{equation}

During the training process, we found that an LR of 0.01 worked well across both tasks. Using Equation \ref{eq:Noise_Scale_SGD}, this would give us an equivalent LR of $2.133 \times 10^{-5}$ when training with $1.35 \times 10^6$ trajectories. We used this value as a baseline and we ended up setting an LR of $2.133 \times 10^{-4}$ for training the final model on a larger data set, as can be seen in Table \ref{tab:hyper_parameters}.


%

\begin{table}[ht]
    \centering
    \begin{tabular}{|l|c|}
		\hline
		Parameter & Value\\
		\hline
		Batch Size & 32\\
		Num. Heads & 16\\
		CNN Dropout & 0.05\\
		Trans. Dropout & 0\\
		Learn Rate & 0.0002133\\
		Num. Epoch & 100\\
		Patience & 10\\
		\hline
    \end{tabular}
    \caption{Table of final hyper-parameters used to train the models for Task 1 and 2.}
    \label{tab:hyper_parameters}
\end{table}

Our testing revealed that using smaller batches and more heads improved performance. However, decreasing the batch size greatly increases the running time. Ideally, we would have used 32 or more heads. However, we were constrained by our equipment's 8GB video memory (GTX 1070). Thus, with the above set of hyper-parameters, we strove to attain a good balance of speed and performance with our hardware constraints.

\subsection{Model Training}
\label{sec:model_training}
    All model training was conducted in Python 3.8.5 using Pytorch in Jupyter Notebook 6.0.3. For simplicity, we trained all our models for Task 1 and Task 2 on data sets of size 2 million using the same split as before: 75\% of the data for training and 25\% for testing, with the training set further broken as 90\% for training and 10\% for validation to be used by Early Stopping to halt the training \cite{EarlyStopping_Bjarten}. Thus, the final training sets were broken up into: $1.35\times10^6$  million trajectories for training, $1.50\times10^5$ for validation, and $5\times10^5$  for testing.\medskip

For Task 1, we trained 12 models, each model corresponding to a batch of trajectory lengths: [10,20], [21,30], [31,40], [41,50], [51,100], [101,200], [201,300], [301,400], [401,500], [501,600], [601,800], and [801,1000]. 
All datasets are of size $2 \times 10^6$, with the aforementioned training/test/validation split ratio. By default, the andi-datasets package \cite{AnDi-package} generates trajectories with anomalous exponent $\alpha \in [0.05, 2)$ in intervals of $0.05$. This means that 39 different alpha values can be generated, and there are five diffusion models for a total of 195 different kinds of trajectories. After generating data sets of size $2 \times 10^6$ would ensure that each of the 195 combinations of diffusion model and $\alpha$ has a representative sample size of about $10^5$. Naturally, this is lower after splitting the data sets into training, test, and validation. However, data sets of this size were a good compromise between model performance and training time on our hardware.\medskip 

In order to accelerate the training of the 12 models, we reduced the patience of our early stopping function from ten, used in hyper-parameter selection, to five while maintaining the number of epochs at 100. Additionally, we conducted the training so that models inherit the parameter state of a previously trained model. This has two advantages: Firstly, it indirectly exposes the model to more unique trajectories, as the model will inherit a parameter state that was trained on a different data set, thus reducing overfitting. Secondly, it jump-starts the training of each model with a parameter state that was trained on longer trajectories, which should contain relevant information for classifying shorter trajectories.\medskip

 To implement this training scheme, we trained the first ConvTransformer on the \emph{easiest} dataset, trajectories of length $[801-1000]$, as can be inferred from our testing and the results in the AnDi Challenge \cite{munoz-gil2021_AnDi_Challenge}. Then the parameter state of this model is used as the starting parameters state for the next model, which will be trained on trajectories of lengths $[601-800]$ and so forth until the final model is trained on trajectories of length $[10,20]$. Once we have completed the first training pass through, we loop back to the top and repeat the process, with model $[801, 1000]$ from round two inheriting the parameter state of model $[10, 20]$ from the first training round. Finally, the round two models are tested on every testing data set, and the best models at each trajectory length are selected. This final selection process resulted in 11 models as the models trained on $[401,500]$, $[601-800]$, and $[501-600]$ outperformed other models at their native trained trajectory lengths. Thus, our compiled model for Task 1 consists of 11 models, each of which is in charge of certain trajectory lengths.\medskip


Finally, for Task 2, a single model was used across all trajectory lengths (10 to 1000) as we were to improve upon the state art while maintaining parsimony, as we show in Figure \ref{tab:ranking_AnDi_Challenge}. The transformer was trained using 100 epochs, patience of 10, and a single data set with trajectory lengths $[10, 1000]$.\medskip

\section{Results}

%

We use the AnDi Interactive Tool\footnote{\url{http://andi-challenge.org/interactive-tool/}} extensively in our testing in order to be able to assess our model's performance against the current state of the art. However, in order to gain further insight into model's performance under different combinations of trajectory type (ATTM, CTRW, LW, FBM, SBM), trajectory length, anomalous diffusion exponent ($\alpha$), and signal to noise ratio (SNR), defined as SNR$= \sigma_{{\rm disp}}/\sigma_{{\rm noise}}$, where $\sigma_{{\rm disp}}$ is the standard deviation of the displacements and $\sigma_{{\rm noise}}$ is the standard deviation of the Gaussian white noise. We generated datasets for all of the permutations seen in Table \ref{tab:testing_datasets_oscar}.\medskip

  	\begin{table}[ht]
    \centering
    \begin{tabular}{|c|c|c|c|}
  			\hline
  			Diff. Model & Traj. Length & SNR & $\alpha$\\ 
  			\hline
  			ATTM & 10, 20,\ldots, 50, 100, 200, \ldots, 600, 800, 1000 & 1,2 & 0.1, 0.2 \ldots 1.0\\
  			CTRW & 10, 20,\ldots, 50, 100, 200, \ldots, 600, 800, 1000 & 1,2 & 0.1, 0.2 \ldots 1.0\\
  			FBM & 10, 20,\ldots, 50, 100, 200, \ldots, 600, 800, 1000 & 1,2 & 0.1, 0.2 \ldots 1.9\\
  			LW & 10, 20,\ldots, 50, 100, 200, \ldots, 600, 800, 1000 & 1,2 & 1.0, 1.1 \ldots 1.9\\
  			SBM & 10, 20,\ldots, 50, 100, 200, \ldots, 600, 800, 1000 & 1,2 & 0.1, 0.2 \ldots 1.9\\
  			\hline
    \end{tabular}
    \caption{A testing dataset of size 2000 was generated for all the permutations of each row in the table.}
    \label{tab:testing_datasets_oscar}
  	\end{table}
  	
We have used the performance of our model on these data sets to make the figures in the following sections. In order to improve model comparability we will use the following metrics of performance:\medskip

\begin{itemize}
    \item The Mean Average Error (MAE) is defined as 
    \begin{equation}
    \frac{1}{N} \sum_{j=1}^{N} |\alpha_{j, {\rm pred}} - {\alpha_{j, {\rm true}}}|,
    \end{equation}where $\alpha_{\cdot,{\rm pred}}$ and $\alpha_{\cdot,{\rm true}}$ are the predicted and true $\alpha$ values respectively.
    \item The F1-Score is the harmonic mean of precision and recall and it is defined as 
    \begin{equation}
    \frac{\mbox{True Pos.}}{\mbox{True Pos.}+\frac{1}{2}(\mbox{False Pos.} + \mbox{False Neg.}).}        
    \end{equation}For our purposes, we have used the micro averaged F1-score, that is biased by class frequencies, as it has been considered in the AnDi Challenge.
\end{itemize}

Using the AnDi Challenge interactive tool, we can see how the ConvTransformer would have performed in the AnDi Challenge in Table \ref{tab:ranking_AnDi_Challenge}. Overall the ConvTransformer would have placed in the middle of the top ten of the AnDi Challenge. However, ConvTransformer shines in classifying short trajectories (Task 2). Here, it outperforms the top three models in length trajectories $[10, 50]$. Impressively, the ConvTransformer manages these results by training on a single data set that is small when compared to the training set used by team UPV-MAT, described in \cite{Garibo2021_AnDi}, whose model came within the margin of error of ours at the classification of short trajectories in one dimension. Team UPV-MAT's model was trained using $4 \times 10^6$ trajectories \cite{Garibo2021_AnDi}, while we used only $1.35 \times 10^6$. This is noteworthy because we observed that during the period of patience, after the final saved parameter state, the ConvTransformer continues to converge to a training loss of zero. This indicates that the network is not fully saturated. Thus, it is possible that given a larger training set, the ConvTransformer would learn, which should lead to an increase in performance.\medskip

  	\begin{table}[ht]
    \centering
    \begin{tabular}{|c|c|c|c|c|}
  			\hline
  			Task & Rank & MAE & Team Name & Method\\ 
  			\hline
  			1 & 7 & 0.453 & ConvTrans. & ConvTransformer\\
  			1 & 1 & 0.326 & UPV-MAT & CNN + biLSTM \cite{Garibo2021_AnDi}\\
  			1 & 2 & 0.329 & HNU & LSTM\ \cite{Li_2021}\\
  			1 & 3 & 0.385 & eduN & RNN + Dense NN \cite{Argun2021ClassificationIA}\\
  			\hline
  			Task & Rank & F1 Score & Team Name & Method\\ 
  			\hline
  			2 & 6 & \textbf{0.563} & ConvTrans. & ConvTransformer\\
  			2 & 1 & 0.499 & eduN & RNN + Dense NN \cite{Argun2021ClassificationIA}\\
  			2 & 2 & 0.560 & UPV-MAT & CNN + biLSTM \cite{Garibo2021_AnDi}\\
  			2 & 3 & 0.525 & FCI & CNN \cite{GRANIK2019185,bai2018empirical}\\
  			\hline
    \end{tabular}
    \caption{Rank is the overall ranking on the entire AnDi Challenge test data set in one dimension. The MAE, and F1-Scores are calculated only for short trajectories, of length 10 to 50 in one dimension, by the AnDi Interactive Tool.}
    \label{tab:ranking_AnDi_Challenge}
  	\end{table}

\subsection{Regression of the anomalous diffusion exponent (Task 1)}

\begin{figure}[h]
	\centering	
	\includegraphics[width=9cm]{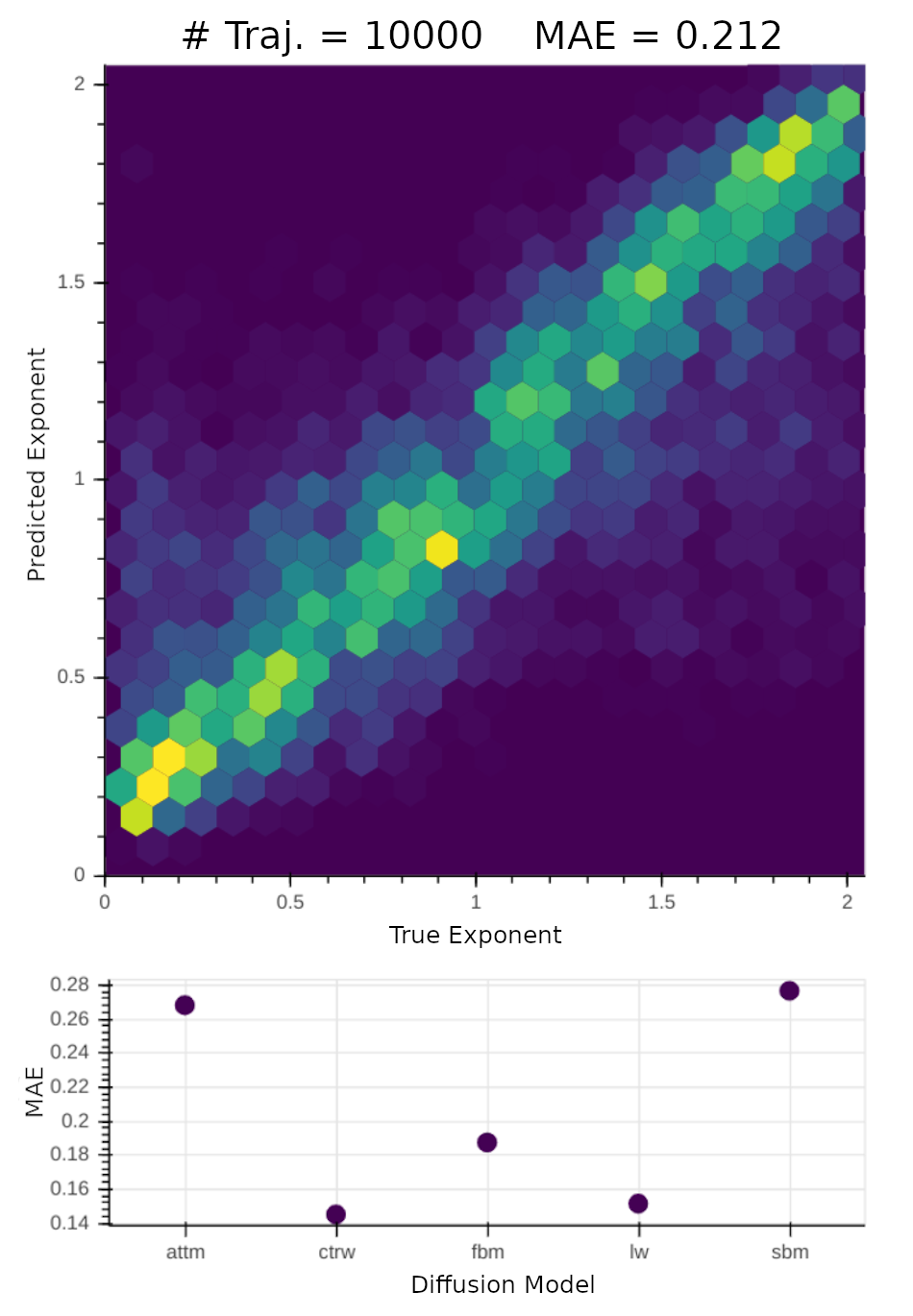}
	\caption{ConvTransformer performance in the regression task measured by the AnDi Interactive tool. Lighter colours represent higher frequencies.}
	\label{fig:task1_performance_AnDi_screenshot}
	\vspace{-1em}
\end{figure}

We first show the performance of our model with the AnDi Interactive tool, see Figure \ref{fig:task1_performance_AnDi_screenshot} .The ConvTransformer, as well as the top performers in the AnDi Challenge seen in Table \ref{tab:ranking_AnDi_Challenge}, had the most difficulty inferring the $\alpha$ of ATTM and SBM diffusive regimes, with ATTM being far more problematic. This makes sense if we consider the way ATTM trajectories are generated. The displacements of particles undergoing ATTM are distributed $BM(D, t, \Delta t)$, where $BM$ generates a Brownian motion trajectory of length $t$ sampled at times $\Delta t$, with diffusivity coefficient $D$. Additionally, in ATTM $D$ is re-sampled every $t \sim D^{\sigma/\alpha}$. This means that every time $t$ a particle in ATTM will change diffusive regime in a manner that may obscure $\alpha$.\medskip

Similar to ATTM, SBM also experiences changes in $D$, the diffusivity coefficient. However, in SBM $D(t) = D\psi(t)$ \cite{dossantos_2021_sbm_diffusivity}. On the surface it would appear as though a gradual change in diffusivity should not appear to pose as much difficulty as the regime shifts in ATTM. However, it may have contributed to the difficulty of inferring $\alpha$ in SBM trajectories.\medskip

\begin{figure}[h]
	\centering	
	\includegraphics[width = 10cm]{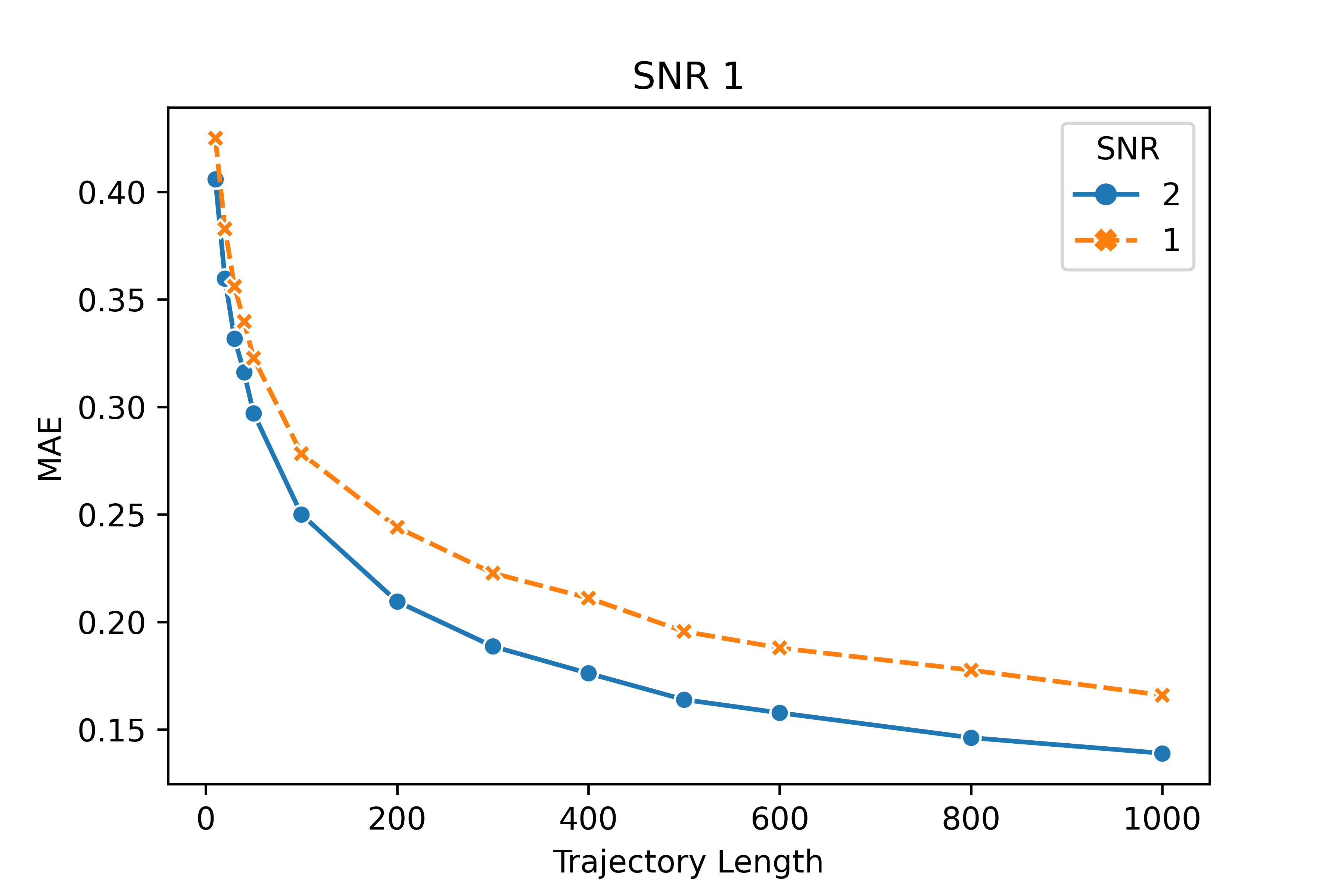}
	\caption{ConvTransformer performance (MAE) in the regression of the anomalous diffusion exponent by SNR as a function of trajectory length.}
	\label{fig:fig2_TLvsMAE_hue_SNR_Task1}
	\vspace{-1em}
\end{figure}

ConvTransformer performance scales as expected with trajectory length, see Figure \ref{fig:fig2_TLvsMAE_hue_SNR_Task1}. That is to say, as trajectory length increases, the model performance also improves. Notably, performance scaled less erratically than in other models, such as the best performing model in Task 1 \cite{Garibo2021_AnDi}. Interestingly, noise does not affect model accuracy as heavily at shorter trajectory lengths, and the performance difference between trajectories with respect to the SNR appears to stabilize after trajectories of length $\sim 200$.\medskip

\begin{figure}[h]
	\centering	
	\includegraphics[width = 17cm]{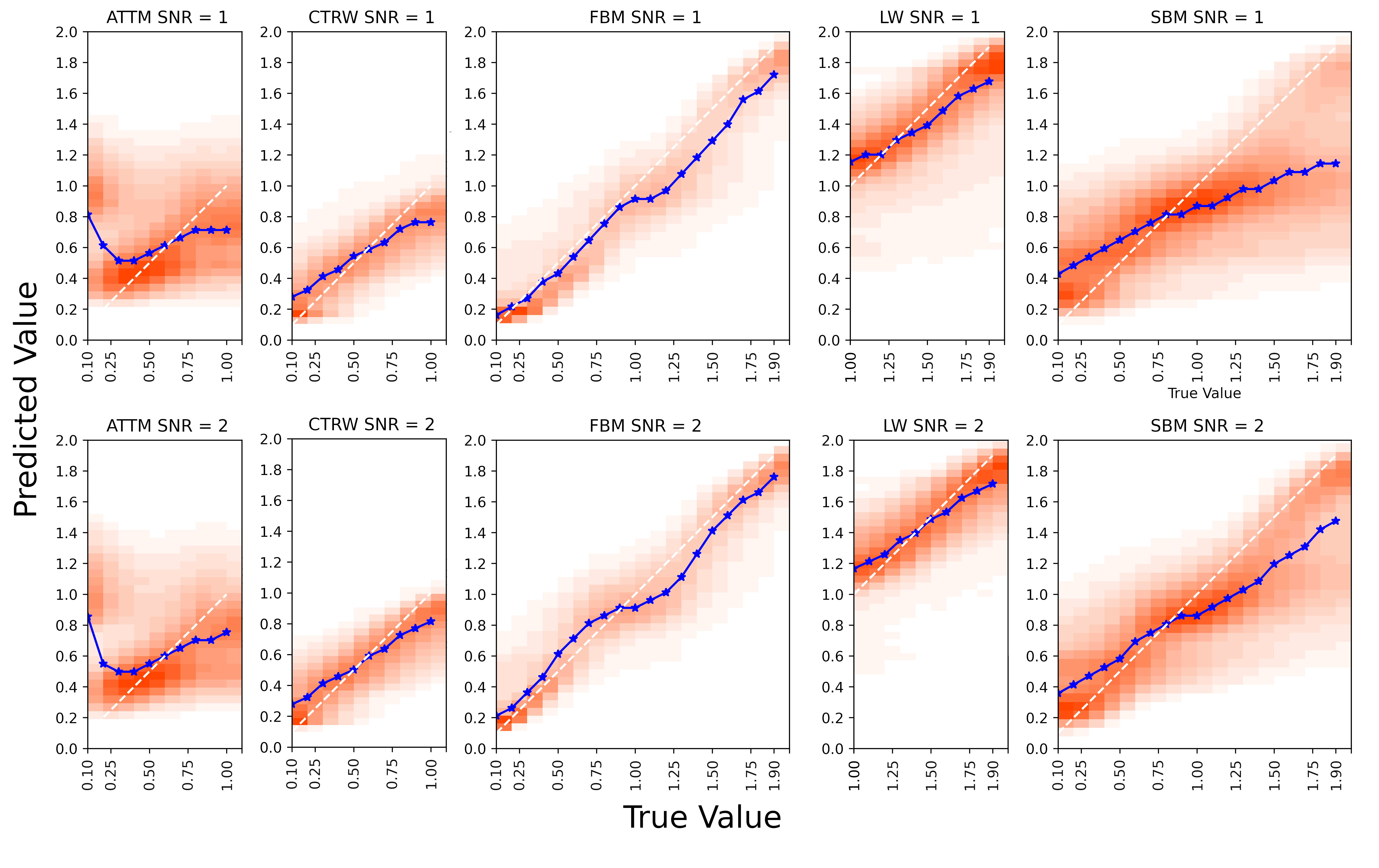}
	\caption{Heat map of ConvTransformer performance in the regression of the anomalous diffusion exponent showing True and Predicted $\alpha$ by the underlying diffusion model. The blue line denotes the median value of the true $\alpha$ values. Predicted values of $\alpha$ are shown from $[0, 2]$. However, there were a few instances where the ConvTransformer predicted values marginally less than 0 and greater than 2.}
	\label{fig:Task1_heatmaps_SNR1and2}
	\vspace{-1em}
\end{figure}

In Figure \ref{fig:Task1_heatmaps_SNR1and2}, we can see the performance breakdown by the underlying diffusive regime. These plots shed more light on the performance issues when regressing $\alpha$ for ATTM. From Figure \ref{fig:Task1_heatmaps_SNR1and2}, it is evident that most of the difficulty with regressing $\alpha$ in ATTM appears to occur in heavy sub-diffusive trajectories at $\alpha \approx 0.1$, regardless of noise. There, we can see a roughly bi-modal distribution with two clusters at $\alpha \approx 0.4$ and $\alpha \approx 0.9$, with the more significant peak at about $0.9$, as shown by the median value line. 
\medskip 

Additionally, the ConvTransformer shows similar confusion patterns in the regression task for the SBM model, where it confuses highly super-diffusive trajectories with, roughly, normal diffusion (Figure \ref{fig:Task1_heatmaps_SNR1and2}). In both of these cases, the regime shift in ATTM and the change in $D$ in SBM could be making heavy anomalous diffusion (both super and sub-diffusion) appear as though it was normal diffusion. However, these effects could also be an artifact of the training data since all diffusive regimes can exhibit normal diffusion, so there will be more trajectories with $\alpha = 1$ than either super or sub-diffusion, or a combination of both effects.\medskip

\begin{figure}[h]
	\centering	
	\includegraphics[width = 7.5cm]{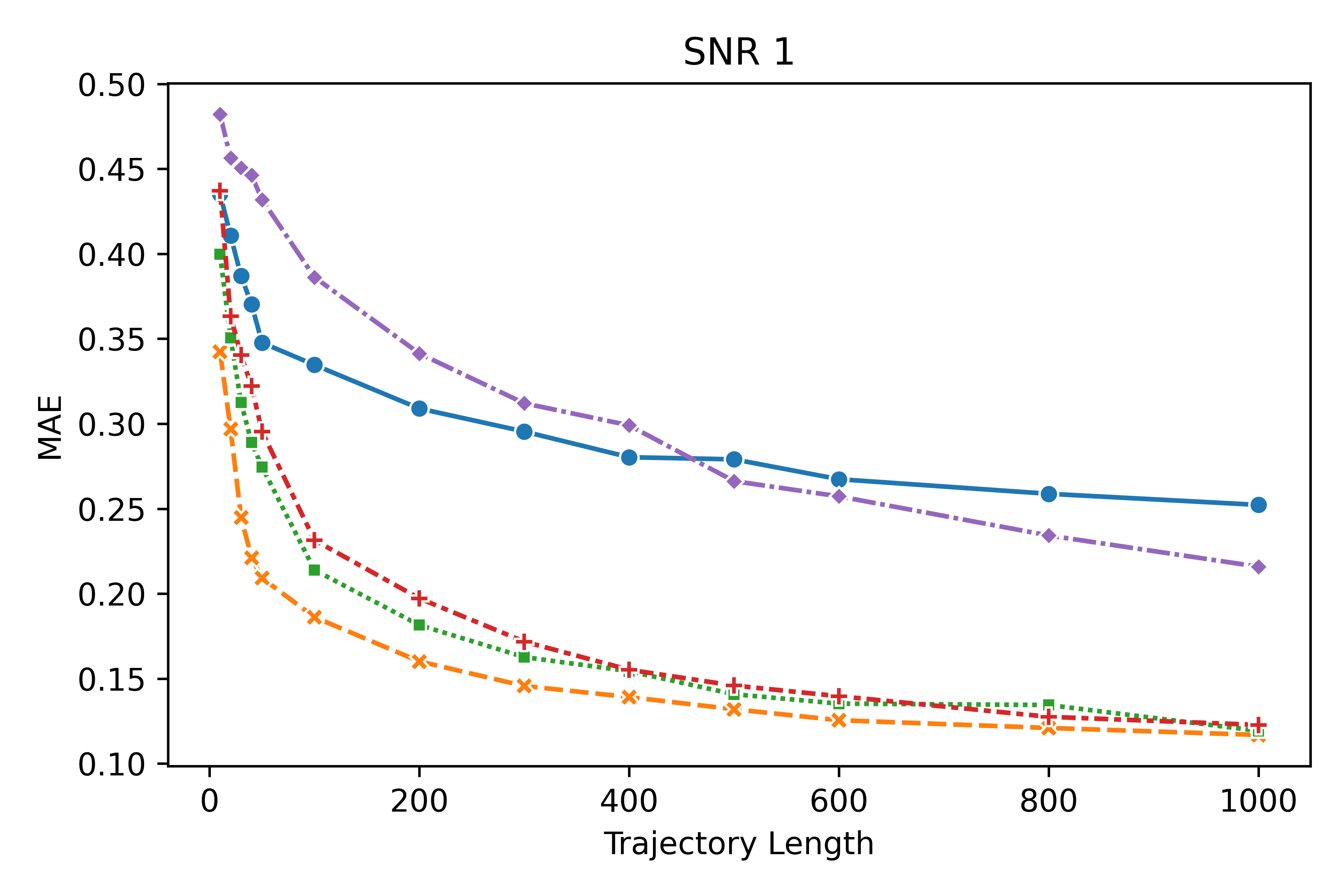}
	\hspace{-1em}
	\includegraphics[width = 7.5cm]{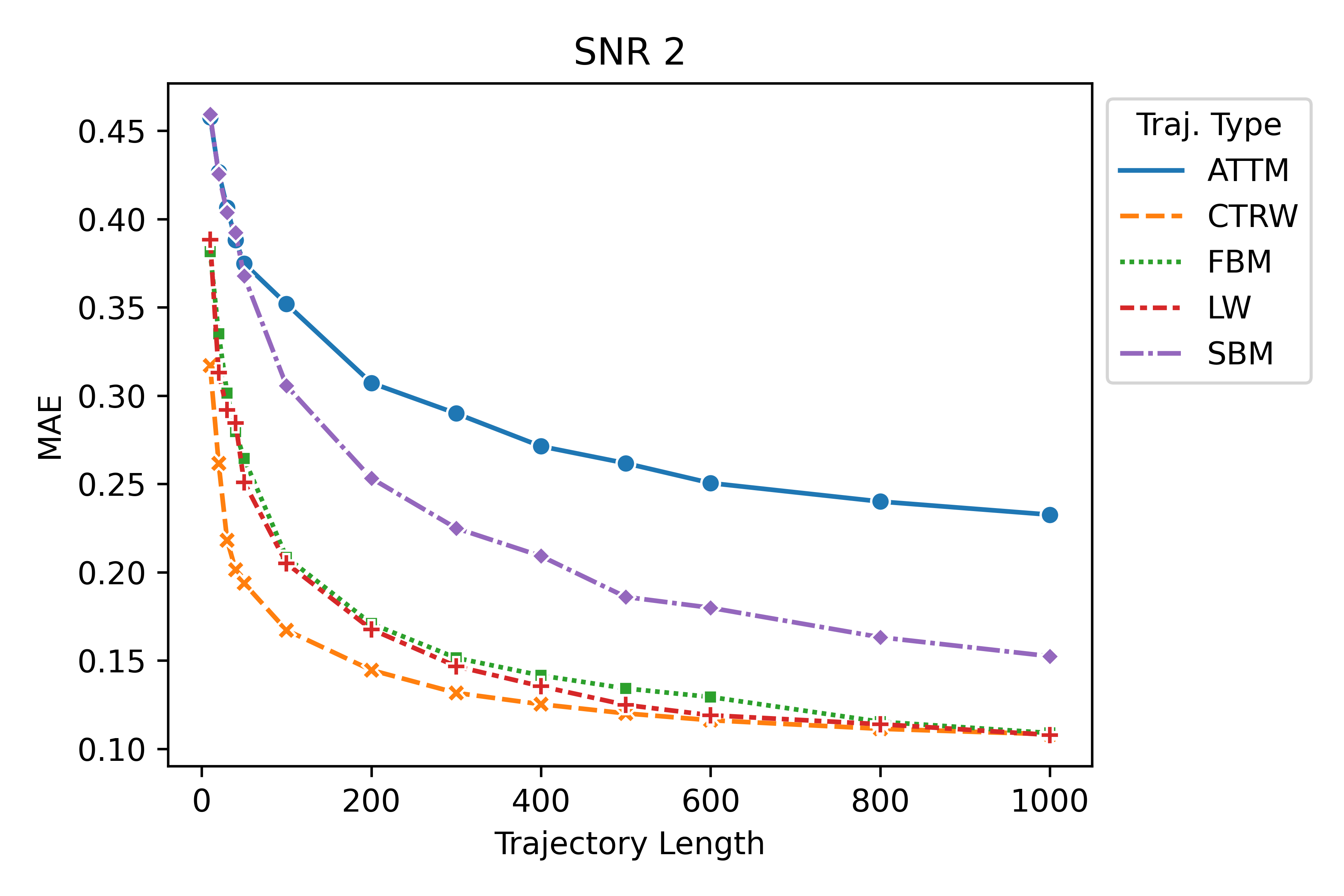}
	\caption{ConvTransformer performance in the regression of the anomalous diffusion exponent (MAE) shown as a function of trajectory length and trajectory type.}
	\label{fig:Fig3_MAEvsTL_hue_trajType}
	\vspace{-1em}
\end{figure}

Figure \ref{fig:Fig3_MAEvsTL_hue_trajType} takes a closer look at model performance by trajectory length and type. Once again, most trajectories, with the exception of the ones generated by the SBM model, perform very similarly at SNR 1 and SNR 2, which shows notably worse performance at SNR 1 on all trajectory lengths. It can be verified via the AnDi Interactive Tool that the same effect occurs in the top three models (UPV-MAT, HNU, eduN) shown in Table \ref{tab:ranking_AnDi_Challenge}, where SBM performance is the most sensitive to additional noise in the trajectory.\medskip

When regressing the anomalous diffusion exponent, the sensitivity of machine learning models to added noise in SBM trajectories warrants further study. Recently, Szarek  \cite{szarek_2021_lstm_sbm_noise} has encountered a similar lack in resiliency to noise using an RNN-based model, like UPV-Mat, HNU, and eduN models. It appears that the difficulty in working with SBM is an inherent characteristic of SBM trajectories, as opposed to the neural network architecture used for inference of $\alpha$. This is further substantiated by our transformer-based method encountering the same problem.\medskip

\begin{figure}[h]
	\centering	
	\includegraphics[width = 7.5cm]{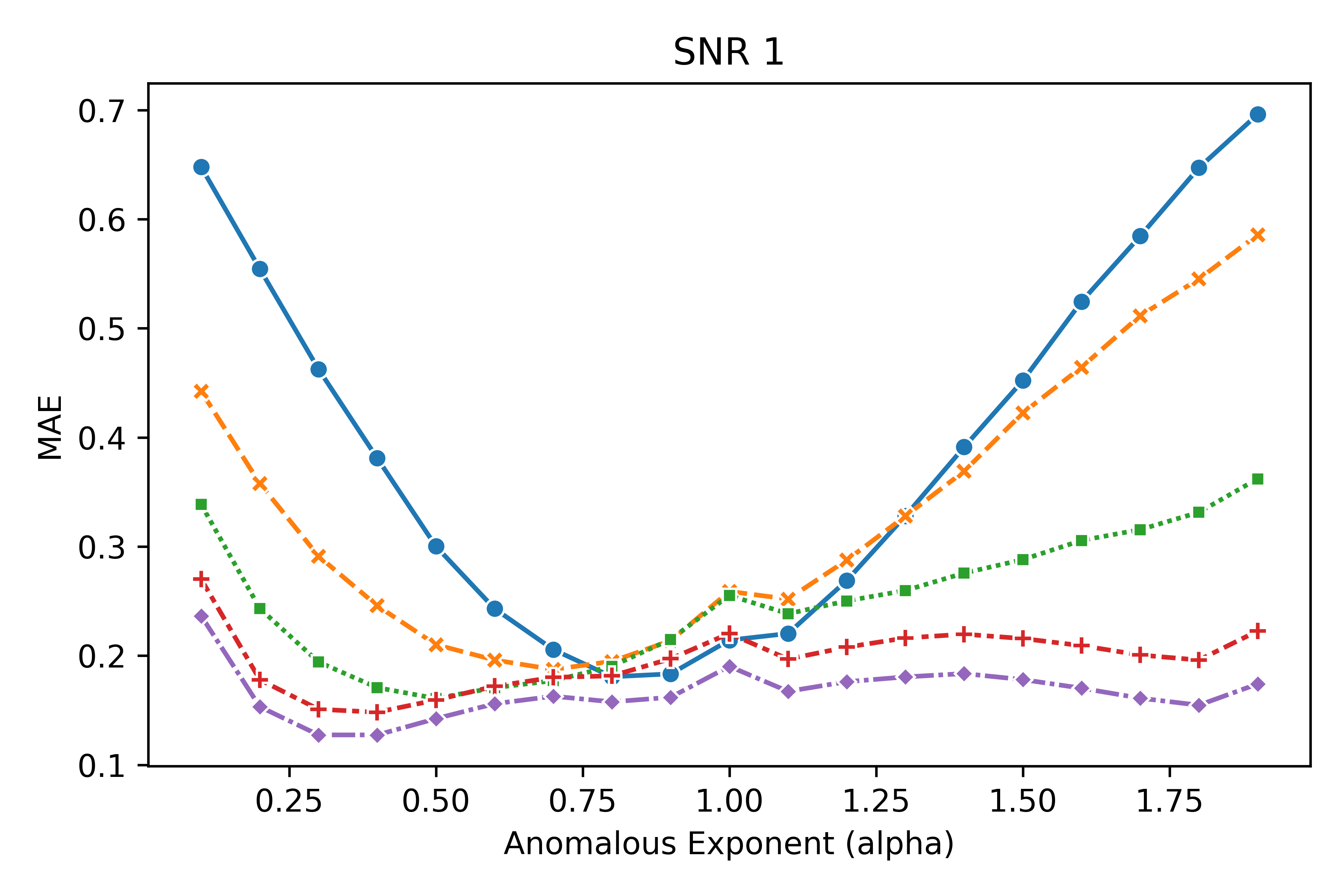}
	\hspace{-1em}
	\includegraphics[width = 7.5cm]{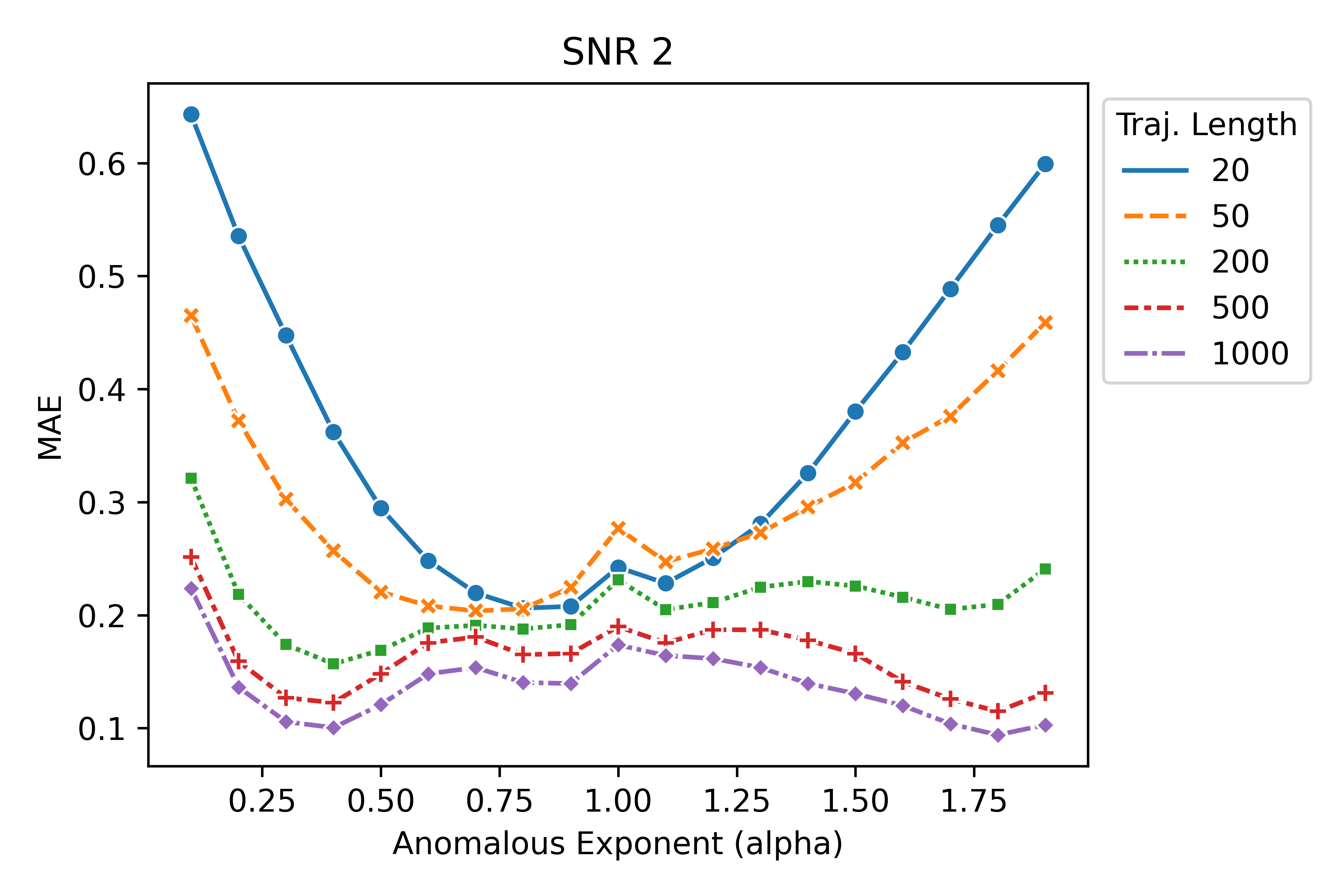}
	\caption{ConvTransformer performance in the regression of the anomalous diffusion exponent (MAE) by trajectory length as a function of $\alpha$, the anomalous diffusion exponent.}
	\label{fig:Fig4_MAEvsAlpha_hue_TrajLength}
	\vspace{-1em}
\end{figure}

\begin{figure}[h]
	\centering	 
	\includegraphics[width = 7.5cm]{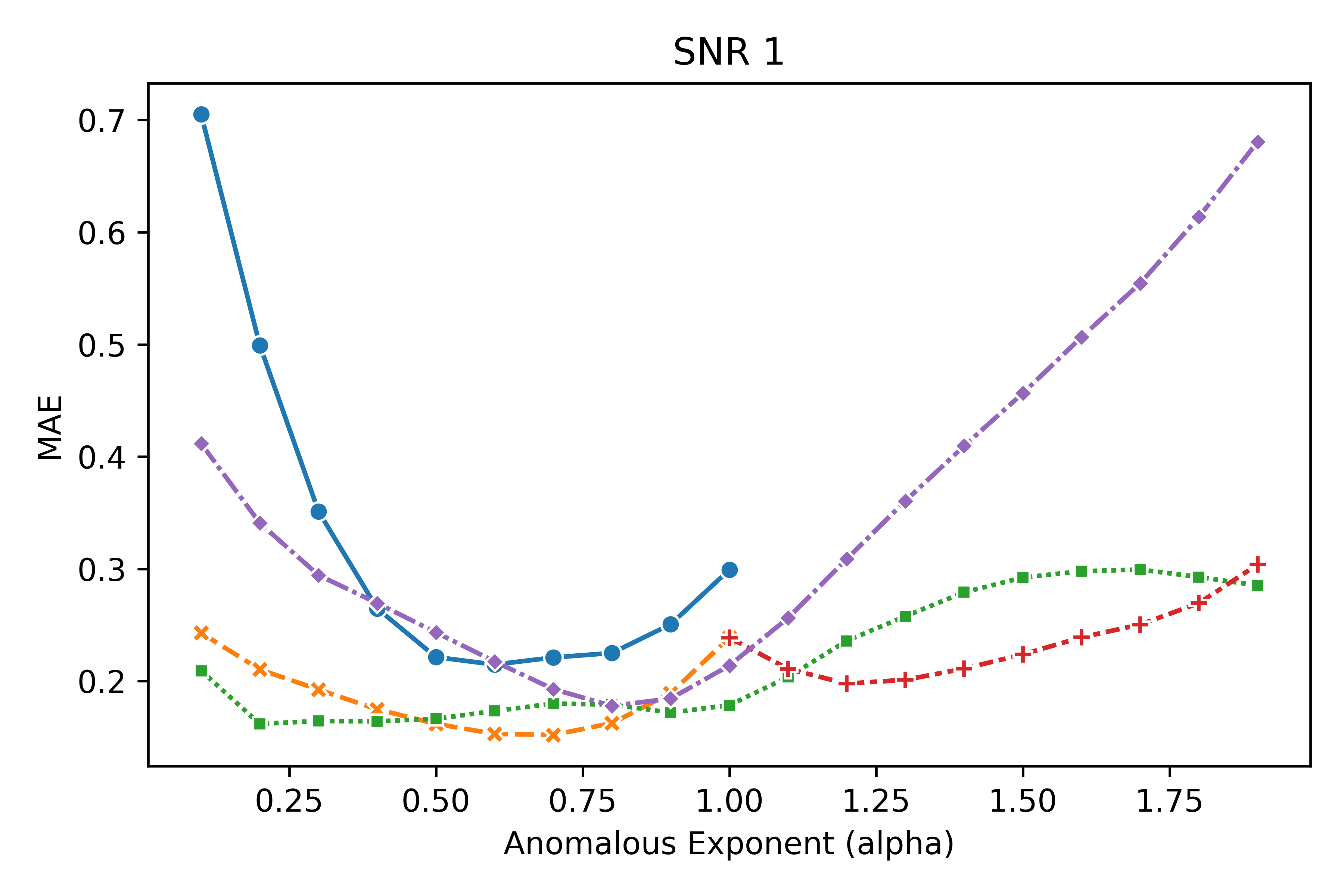}
	\hspace{-1em}
	\includegraphics[width = 7.5cm]{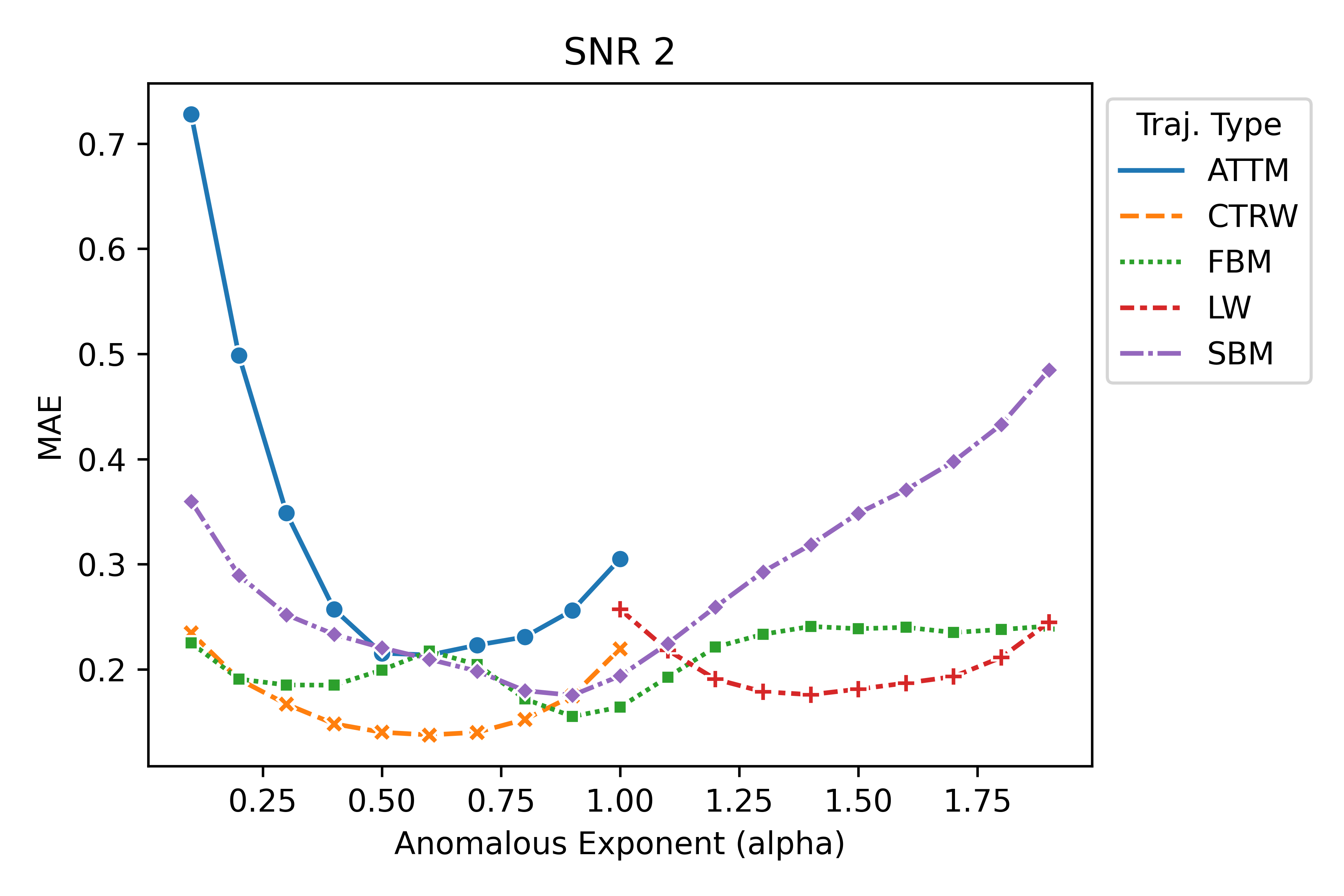}
	\caption{ConvTransformer performance in the regression of the anomalous diffusion exponent (MAE) by underlying diffusive regime as a function of $\alpha$, the anomalous diffusion exponent.}
	\label{fig:Fig7_MAEvsAlpha_hue_trajType}
	\vspace{-1em}
\end{figure}

In terms of model performance for different values of $\alpha$, the ConvTransformer perform best at $\alpha \approx .9$ (Figure \ref{fig:Fig4_MAEvsAlpha_hue_TrajLength}). However, for long trajectories, those with 200 or more points, the model performs best roughly between $\alpha \in [0.25, 0.5]$. The latter scenario seems to be closer to the truth if we examine the model performance at various levels of $\alpha$ by trajectory type as in Figure \ref{fig:Fig7_MAEvsAlpha_hue_trajType}. Our ConvTransformer performs best roughly in the middle of the domain of $\alpha$ of each trajectory type, with an adequate, though not optimal, performance at $\alpha \approx 1$ across all trajectory types (Figure \ref{fig:Fig7_MAEvsAlpha_hue_trajType}). Part of the reason performance is overestimated, at $\alpha \approx 1$, when pooling the trajectory types may be that CTRW and SBM perform best at $\alpha \approx 1$, and these two types of diffusion can be super and sub-diffusive. Thus, they have more testing points and skew the pooled values in Figure \ref{fig:Fig4_MAEvsAlpha_hue_TrajLength}.\medskip

\subsection{Classification of trajectories according to the anomalous diffusion generating model (Task 2)}

\begin{figure}[h]
	\centering	
	\includegraphics[width = 14cm]{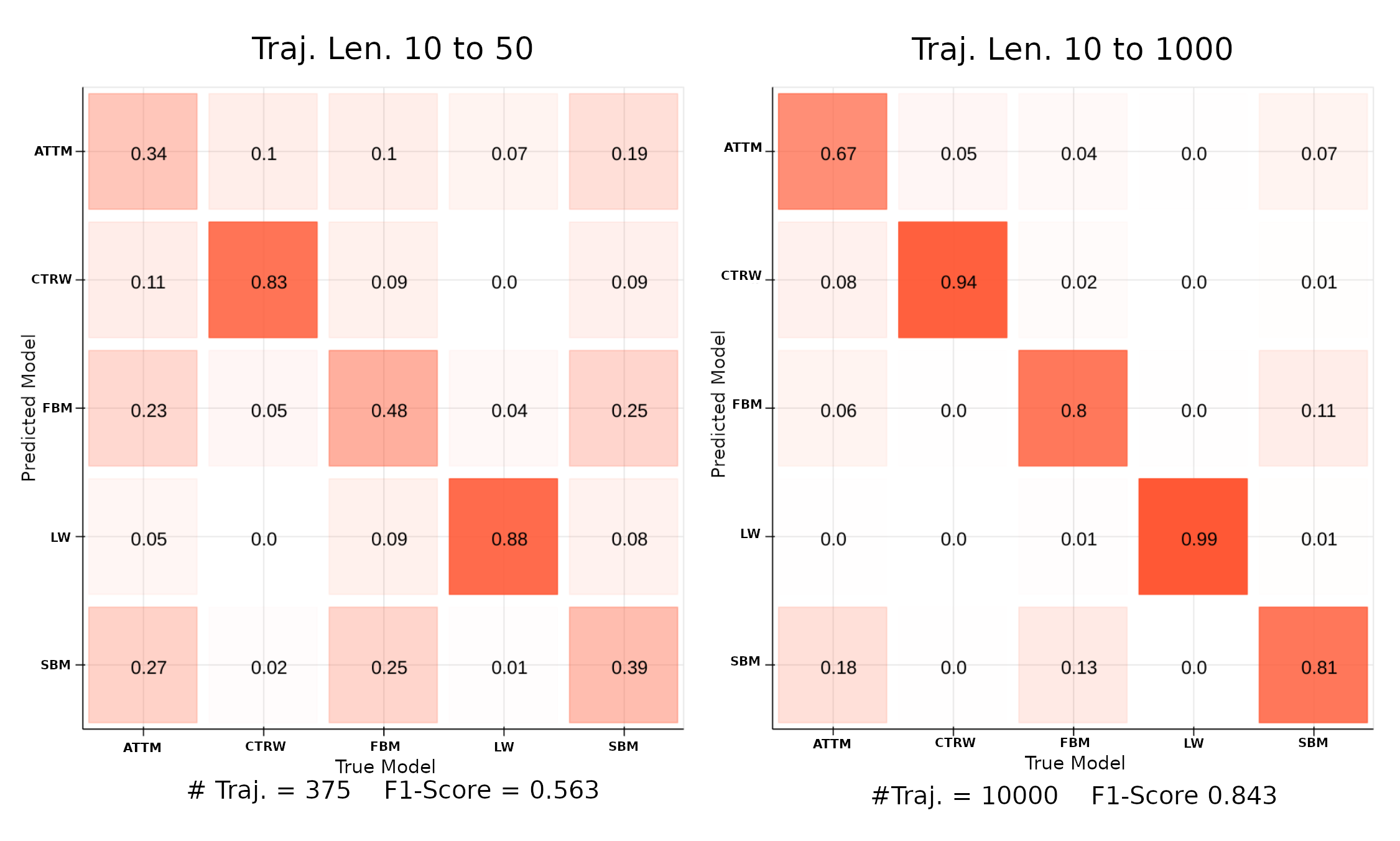}
	\caption{Confusion matrices of ConvTransformer trajectory classification accuracy (Task 2) obtained from the AnDi Interactive Tool.
	}
	\label{fig:task2_andi_interactive}
	\vspace{-1em}
\end{figure}

ConvTransformer performance in the classification of trajectories according to the anomalous diffusion generating model (Task 2) presents results in the average overall, with respect to the ten best models of the AnDi Challenge \cite{munoz-gil2021_AnDi_Challenge}. However, as we mentioned earlier, our ConvTransformer shines in short trajectories. As with the inference of $\alpha$ (Task 1), ATTM trajectories proved to be the most difficult to work with. These trajectories were most often confused with SBM (Figures \ref{fig:task2_andi_interactive} and \ref{fig:task2_confusion_matrices_by_length}). As we have said, this may be because both models have changes in the diffusivity coefficient $D$. If we imagine a short ATTM trajectory, where $D$ only changes a few times, the diffusivity coefficient can increase with time ($D \sim \Phi(t)$) in a way that ATTM could mimic SBM.\medskip

\begin{figure}[h]
	\centering	
	\includegraphics[width = 11cm]{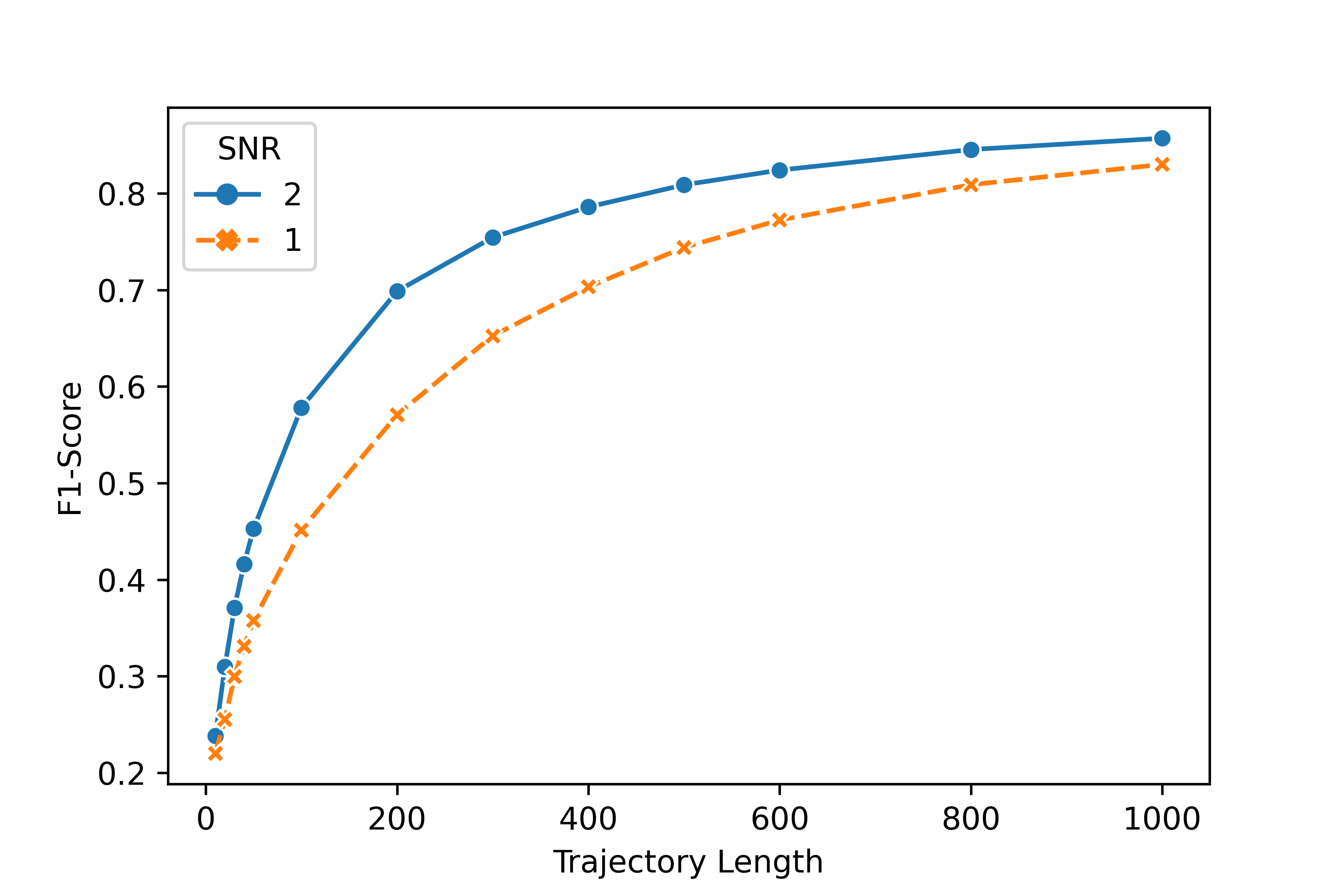}
	\caption{ConvTransformer trajectory classification accuracy (F1-Score) by SNR as a function of trajectory length}
	\label{fig:Fig9_F1vsTL_hue_SNR}
	\vspace{-1em}
\end{figure}

Noise affects ConvTransformer the least at both short and long trajectory lengths (Figure \ref{fig:Fig9_F1vsTL_hue_SNR}). Performance at the lower noise data improves faster with respect to trajectory length. The most significant difference between the two curves, in Figure \ref{fig:Fig9_F1vsTL_hue_SNR}, occurs at trajectories of length $\sim 200$, after which the SNR 1 curve converges towards SNR 2. This indicates that longer trajectories are most helpful when dealing with noisy trajectories that are roughly 200 to 600 dispersals in length.\medskip

When looking at F1-score by the underlying diffusion model, we can see that ConvTransformer performance varies significantly across our five diffusive regimes (Figure \ref{fig:Fig10_F1vsTL_Hue_TrajType}). That being said, unlike Task 1, performance change with respect to noise remains fairly constant across the different kinds of diffusion. The outlier to this behavior is short CTRW trajectories at SNR 1 (Figure \ref{fig:Fig10_F1vsTL_Hue_TrajType}). In this case, model performance is better for the shortest trajectories and then drops off before resuming the expected convergence behavior of F1-Score with respect to the trajectory length. The cause of this artifact in the F1-Score is that at SNR 1 and short trajectory lengths ($[10, 50]$), the ConvTransformer is inclined to classify the other diffusive regimes, with the exception of LW, as CTRW (Figure \ref{fig:task2_confusion_matrices_by_length}). It is noteworthy that the ConvTransformer can make the distinction between LW and CTRW as LW can be considered a special case of CTRW \cite{munoz-gil2021_AnDi_Challenge}.\medskip

\begin{figure}[h]
	\centering	
	\includegraphics[width=15cm]{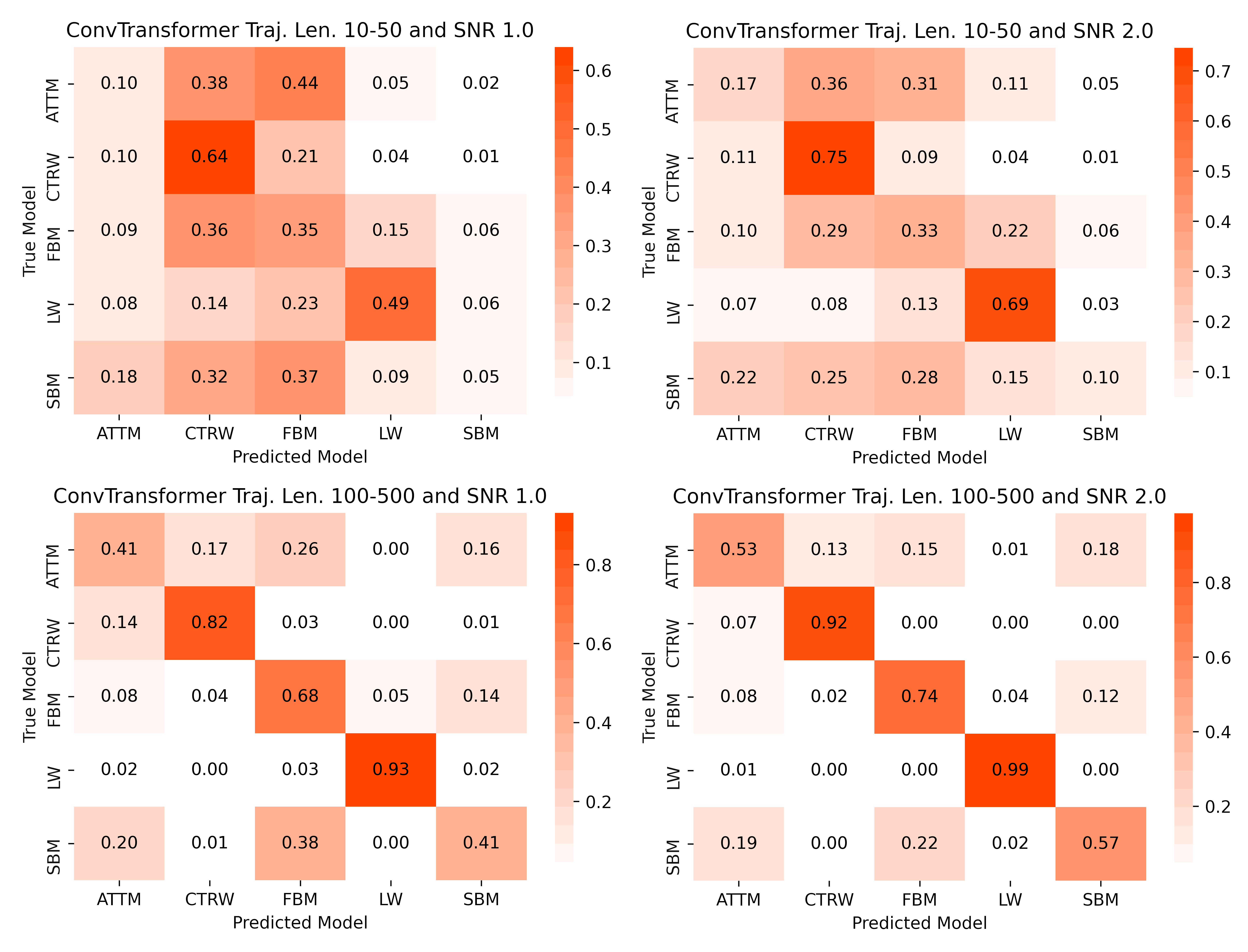} 
	\caption{Confusion matrices showing ConvTransformer classification accuracy (Task 2) at different noise levels. Trajectories of length greater than 500 were omitted because although model performance improves at these lengths it does so as we would expect from Figure \ref{fig:Fig9_F1vsTL_hue_SNR} and does not provides further information.}
	\label{fig:task2_confusion_matrices_by_length}
	\vspace{-1em}
\end{figure}

In terms of ConvTransformer performance in classification (Task 2) with regards to $\alpha$, we can see that ConvTransformer performs better at a value of $\alpha \approx 0.5$ and at the higher-end $\alpha \geq 1.5$, with an apparent plateauing behavior at the upper end of the $\alpha$ domain in longer trajectories with lower noise (Figure \ref{fig:Fig10_F1vsalpha_hue_TrajLen}). In Figure \ref{fig:Fig7_F1vsAlpha_hue_TrajType} we again look at F1-Score as a function of $\alpha$. However, this time we look at the relationship in terms of the underlying diffusive model. Most diffusive models retain the relationship seen in Figure \ref{fig:Fig10_F1vsalpha_hue_TrajLen}, within their respective domains of $\alpha$. However, CTRW and LW deviate from this behavior. Both CTRW and LW appear to have a more linear relationship between F1-Score and $\alpha$, with CTRW performing best at low values of $\alpha$ and LW performing best at higher values of $\alpha$. This relationship strength (F1-Score $\sim \alpha$)  appears to be exacerbated by noise.\medskip

\begin{figure}[h]
	\centering	
	\includegraphics[width = 7.5cm]{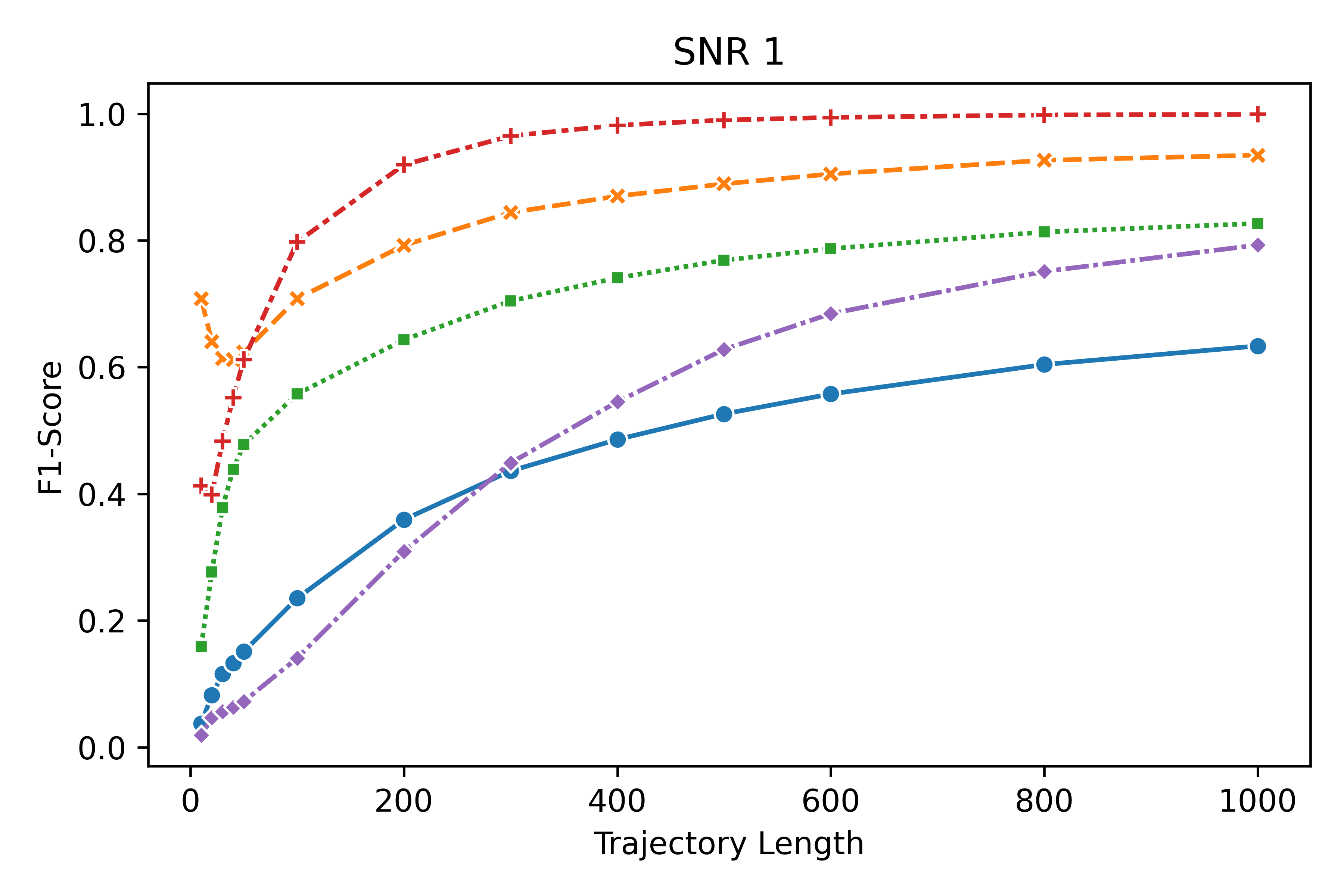}
	\hspace{-1em}
	\includegraphics[width = 7.5cm]{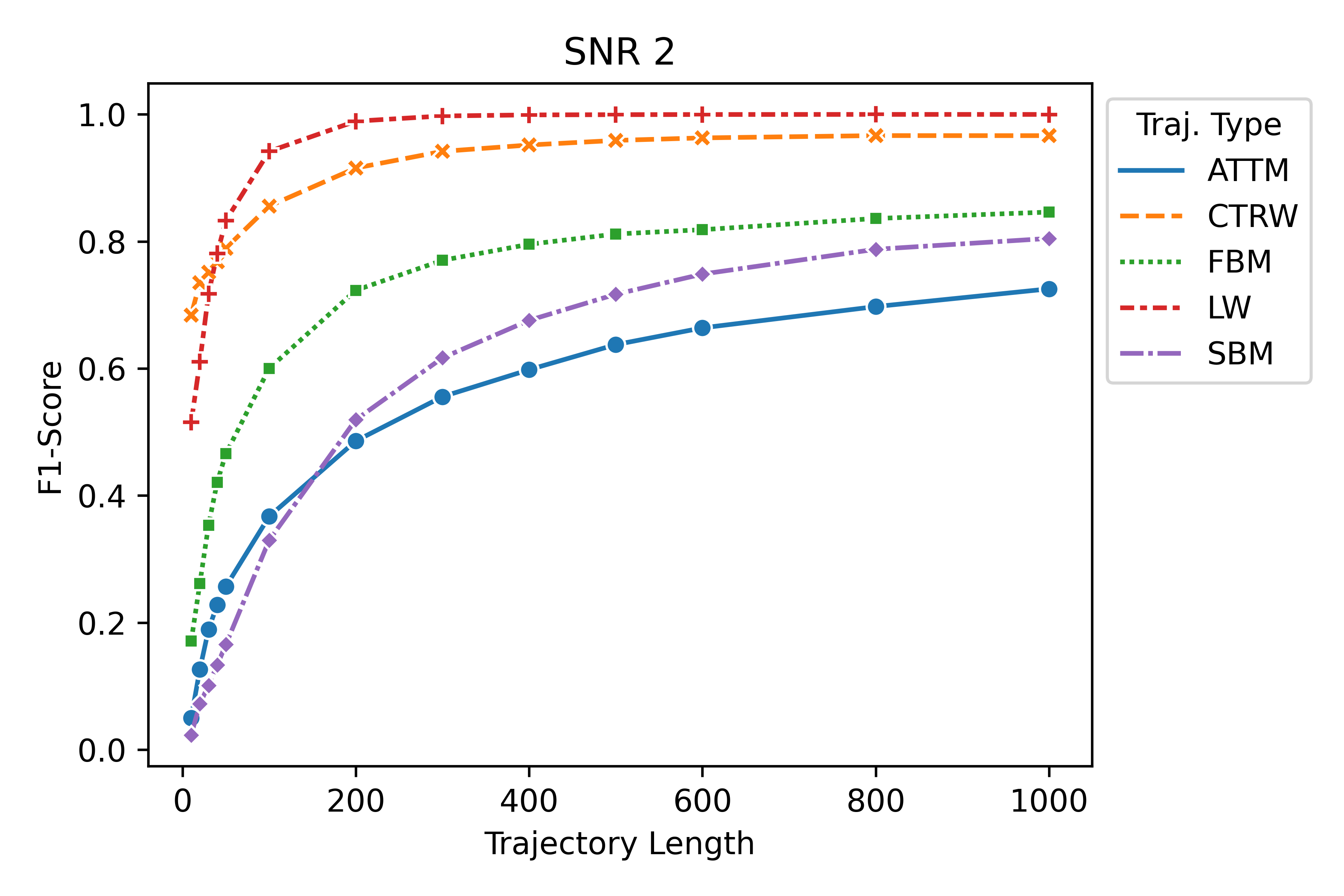}
	\caption{ConvTransformer trajectory classification accuracy (F1-Score) as a function of the trajectory length.}
	\label{fig:Fig10_F1vsTL_Hue_TrajType}
	\vspace{-1em}
\end{figure}

\begin{figure}[h]
	\centering	
	\includegraphics[width = 7.5cm]{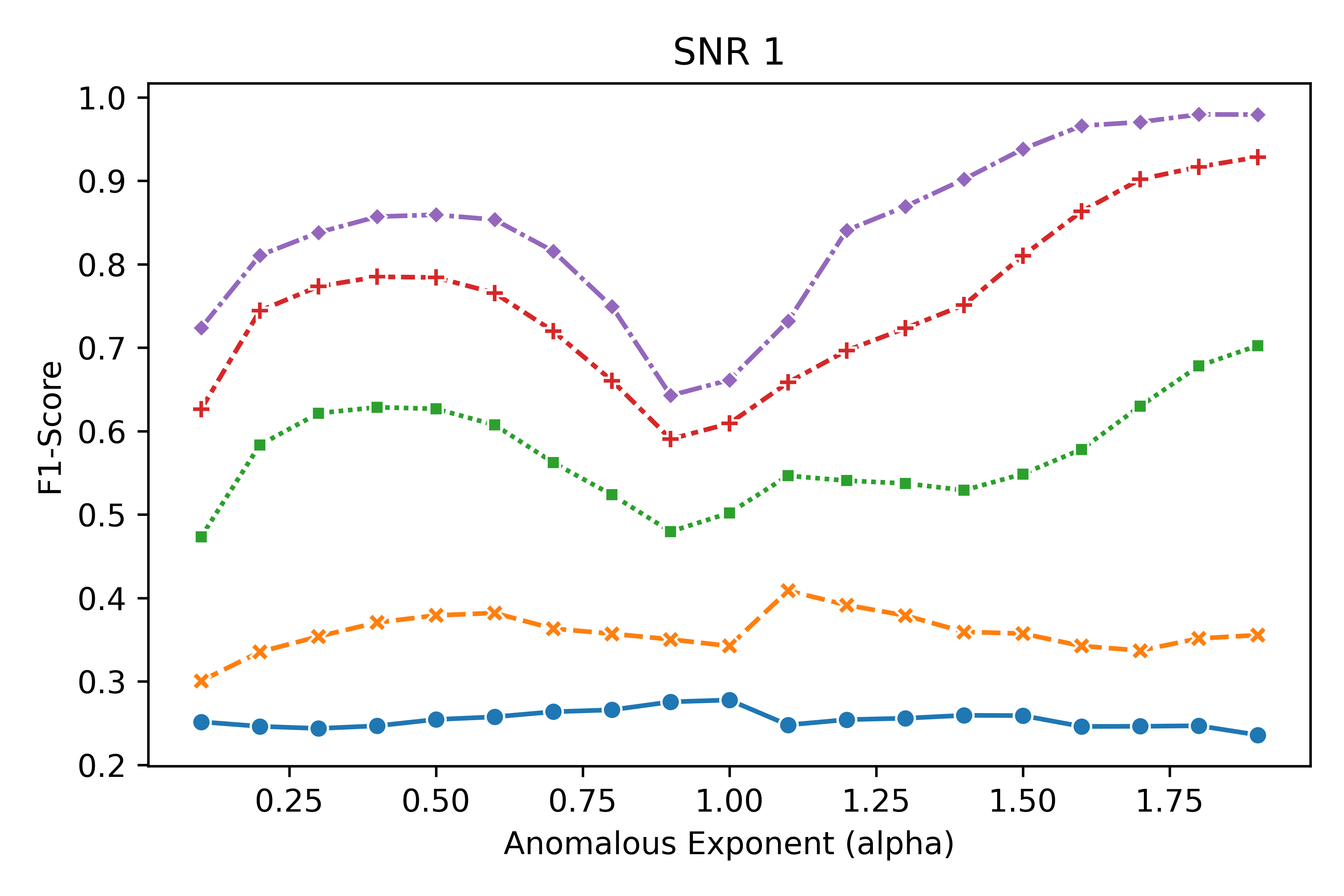}
	\hspace{-1em}
	\includegraphics[width = 7.5cm]{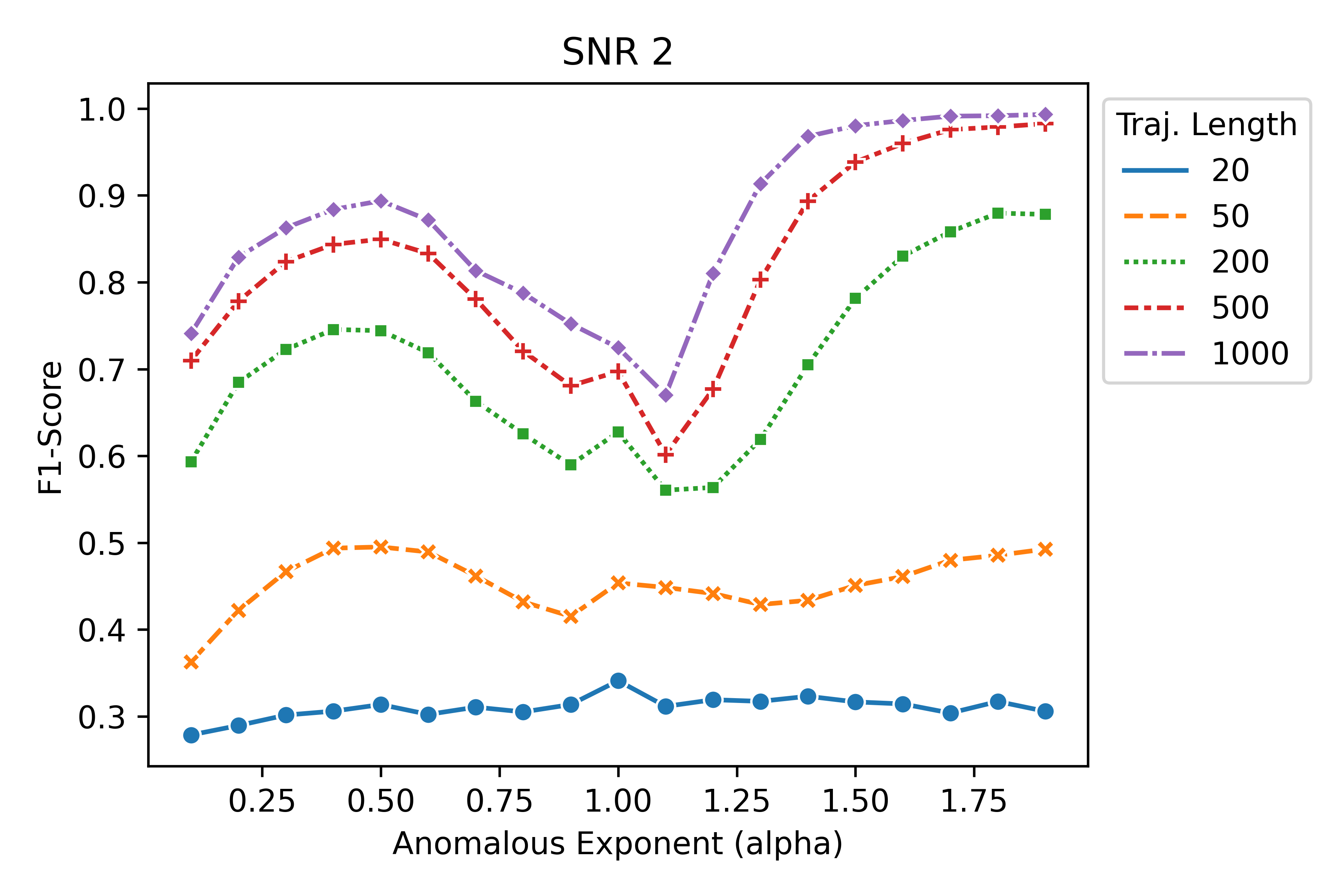}
	\caption{ConvTransformer trajectory classification accuracy (F1-Score) by trajectory length as a function of the anomalous diffusion exponent ($\alpha$).}
	\label{fig:Fig10_F1vsalpha_hue_TrajLen}
	\vspace{-1em}
\end{figure}

\begin{figure}[h]
	\centering	
	\includegraphics[width = 7.5cm]{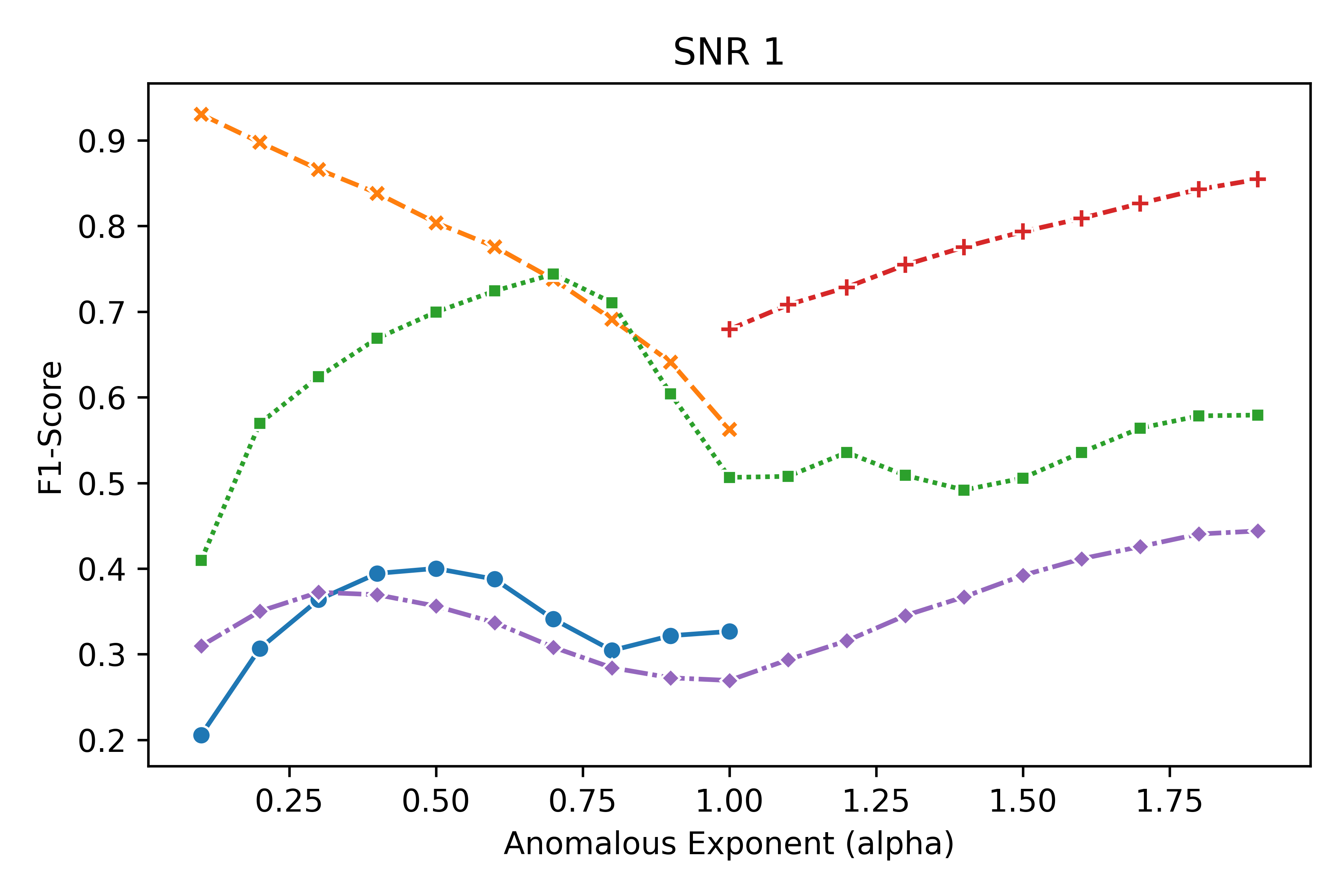}
	\hspace{-1em}
	\includegraphics[width = 7.5cm]{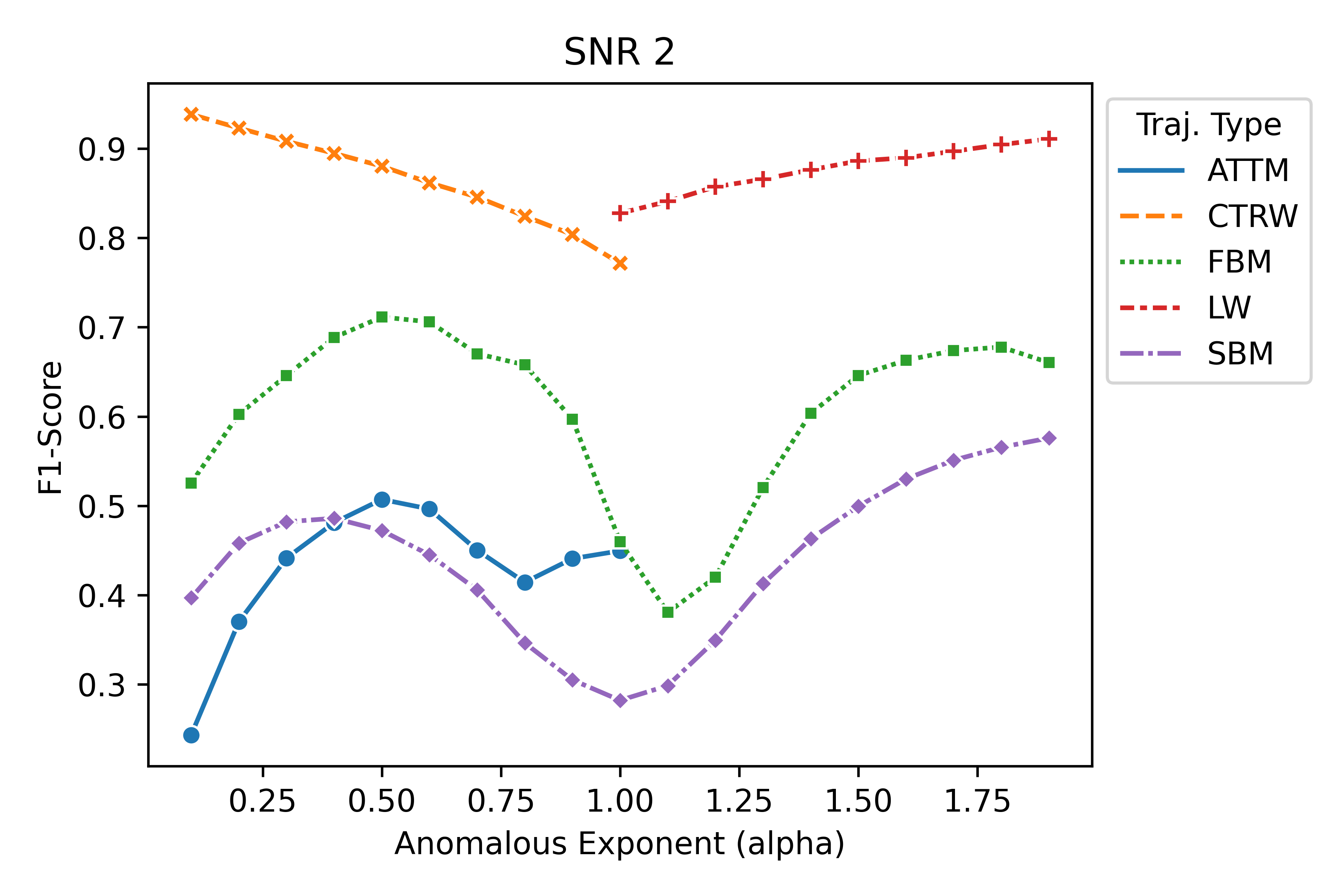}
	\caption{ConvTransformer trajectory classification accuracy (F1-Score) by underlying diffusive regime as a function of the anomalous diffusion exponent ($\alpha$).}
	\label{fig:Fig7_F1vsAlpha_hue_TrajType}
	\vspace{-1em}
\end{figure}

\section{Conclusions}
The primary purpose of this paper was to introduce our new architecture, the ConvTransformer, for the analysis of anomalous diffusion trajectories. To the best of our knowledge, this is the first transformer based architecture to characterize anomalous diffusion. Indeed it is only recently that anyone else has produced a convolutional transformer (for computer vision) \cite{guo2021_CMT_computer_vision_meets_transformer, Liu2020_ConvTransformer_computer_vision}, with the development of their models being concurrent with ours. However, our ConvTransformer stands out in that it does not use positional encoding and only uses the transformer encoding block from \cite{vaswani2017attention}. As such, it is simpler and easier to implement while still providing state-of-the-art results in trajectory classification (Task 2) in short and noisy trajectories.\medskip\medskip

Inspired by the success of transformers in NLP we set out to replace the recurrent bidirectional LSTM part of the architecture in \cite{Garibo2021_AnDi} by transformers. We have improved the classification of short trajectories accuracy with a model that is trained pretty fast since it can be trained in parallel. When we first started working on this model, there was no native support in PyTorch for transformers. However, when writing this manuscript, transformer encoders and decoders are natively supported. As such, we expect further improvements as ease of implementation and optimized code will lead to more accessibility. This should lead to improved iterations of the model and a finer hyperparametrization tuning. Additionally, the increased optimization and access to newer hardware should increase our ConvTransformers performance for improved usability in experimental research.\medskip

Apart from the direct practical implementation of our model in experimental research, going forwards, we would also like to focus on model interpretability. One of the issues plaguing deep learning is the black box effect. When looking at models, we are often only interested in what we can predict or characterize and tend to overlook what we can learn from parameter weighting. Traditionally, parameter weight would allow us to see simple relation between the input features and our desired prediction. For example, birth weight is a strong predictor of adult height \cite{sorensen1999_BirthWeight}. Furthermore, with traditional models like regression, parameter selection leads to discarding information which also informs us about the features that are not relevant to our subject of study. With the rise of deep learning models, we are no longer looking at features, but rather we ingest the data directly and allow our models to discern these features for themselves. With the exception of deep learning models that use feature engineering, as we saw with group UCL and their CONDOR model \cite{Gentili2021_Condor_AnDi_Volpe}. The naive approach to modeling brought about by ML means that we not only lose all information about features, but we also do not know what features are important.\medskip

As we know from Clark \emph{et al.} \cite{clark2019-what-does-bert-look-at}, in the context of NLP, transformer attention heads tend to focus on specific aspects of syntax. For instance, some attention heads may focus entirely on the next token, while others may attend almost entirely to the periods or breaks in a sentence. Following this logic, it is highly likely that some of our ConvTransformer attention heads are specializing on specific features of the trajectories. Hence, a transformer based architecture could be used to determine what trajectory features are important. In this manner, we could recover some model interpretability, and learn from machine learn models in a similar way to how we have traditionally learned from regression.\medskip



\section*{Bibliography}
\bibliographystyle{plain}
\bibliography{references.bib}

\end{document}